\documentclass[12pt]{article}
\usepackage{graphicx}
\usepackage{amsmath}
\usepackage{authblk}

\providecommand{\keywords}[1]{\textbf{\textit{Keywords---}} #1}

\title{Input Guided Multiple Deconstruction Single Reconstruction neural network models for Matrix Factorization}
\author[1*]{Prasun Dutta}
\author[2]{Rajat K.De}
\affil[1]{Institute of Data Engineering, Analytics and Science Foundation, Technology Innovation Hub, Indian Statistical Institute, 203, B. T. Road, Kolkata, 700108, India}
\affil[2]{Machine Intelligence Unit, Indian Statistical Institute, 203, B. T. Road, Kolkata, 700108, India}
\date{}

\begin{document}

\maketitle

\begin{abstract}
Referring back to the original text in the course of hierarchical learning is a common human trait that ensures the right direction of learning. The models developed based on the concept of Non-negative Matrix Factorization (NMF), in this paper are inspired by this idea. They aim to deal with high-dimensional data by discovering its low rank approximation by determining a unique pair of factor matrices. The model, named Input Guided Multiple Deconstruction Single Reconstruction neural network for Non-negative Matrix Factorization (IG-MDSR-NMF), ensures the non-negativity constraints of both factors. Whereas Input Guided Multiple Deconstruction Single Reconstruction neural network for Relaxed Non-negative Matrix Factorization (IG-MDSR-RNMF) introduces a novel idea of factorization with only the basis matrix adhering to the non-negativity criteria. This relaxed version helps the model to learn more enriched low dimensional embedding of the original data matrix. The competency of preserving the local structure of data in its low rank embedding produced by both the models has been appropriately verified. The superiority of low dimensional embedding over that of the original data justifying the need for dimension reduction has been established. The primacy of both the models has also been validated by comparing their performances separately with that of nine other established dimension reduction algorithms on five popular datasets. Moreover, computational complexity of the models and convergence analysis have also been presented testifying to the supremacy of the models.
\end{abstract}


\section{Introduction}
\label{introduction_IG-MDSR-NMF}
Non-negative Matrix Factorization (NMF) \cite{lee1999learning, lee2001algorithms} breaks down a non-negative matrix into two components based on the desired reduced dimension while maintaining non-negativity. As the output coefficients are all positive, each data point can be represented as the sum of vectors in the basis multiplied by specific coefficients. The property of generating the additive parts-based representation distinguishes NMF from other dimension reduction techniques.

Nowadays, low rank approximation/dimensionality reduction is one of the most popular solutions to deal with the `curse of dimensionality' problem. Dimension reduction with traditional iterative procedures requires manual interventions. Deep learning is a contemporary machine learning technology that learns and accumulates knowledge by experience, without the bottleneck of manual interventions. The basic NMF algorithm is a nice example of an iterative technique.

The semi-NMF model, developed by Trigeorgis et al. \cite{trigeorgis2014deep}, learns the appropriate representations for clustering as well as the attribute hierarchy of the dataset. Additionally, two versions of the method have been developed - Deep WSF \cite{trigeorgis2016deep}, a semi-supervised version, and Deep Semi-NMF (DS-NMF) \cite{trigeorgis2016deep}, a deep neural network-based form of semi-NMF. The models have been applied to image datasets. 

Ye et al. have resolved the community detection problem using the Deep Autoencoder-like NMF (DANMF) model \cite{ye2018deep}. A layer-wise feature learning method based on stacked NMF layers has been presented by Song et al. \cite{song2015hierarchical}. The model has been demonstrated with an application on document categorization. Guo et al. have applied sparsity constraints in a multilayer NMF model to learn localised features on images \cite{guo2019sparse}.

A deep autoencoder-like architecture, called Nonsmooth Nonnegative Matrix Factorization (nsNMF), can learn both part-based and hierarchical features \cite{yu2018learning}. Yang et al. have developed DAutoED-ONMF, a deep autoencoder network for Orthogonal NMF \cite{yang2021orthogonal}. To identify patterns in complicated data, Zhao et al. have devised a deep NMF model that learns underlying basic images \cite{zhao2019deep}. Yu et al. \cite{yu2018learning} and Zhao et al. \cite{zhao2019deep} have shown application on image datasets, whereas, Yang et al. \cite{yang2021orthogonal} have demonstrated the applicability of their model in textual and network datasets in addition to the image dataset.

Graph regularisation has been used by the Deep Grouped NMF (DGNMF) model \cite{zhan2021deep} to learn multi-level features. Using graph regularizers, Deep Semi-NMF-EP effectively represents high-dimensional data maintaining data elasticity \cite{shu2021deep}. DGNMF and Deep Semi-NMF-EP models have been applied to image datasets. Lee and Pang have designed an NMF-based feature extraction method integrated into CNN \cite{lee2020feature}. They have considered acoustic signals as their application area.

An NMF system embedding, a stacked autoencoder, has been used for task-specific nonlinear dimension reduction \cite{zhang2016nonlinear}. Tong et al. have designed the objective function integrating central and global loss functions of the soft label constraint matrix to accomplish dimension reduction \cite{tong2019deep}. Both the studies have used image datasets for their applications. Shu et al. \cite{shu2022robust} have presented Robust Graph regularised Non-Negative Matrix Factorization with Dissimilarity and Similarity constraints (RGNMF-DS), a novel scRNA-seq data representation method for scRNA-seq data clustering.

In our earlier works, we have fused the advantages of conventional iterative learning of NMF with that of deep learning. A shallow neural network model, named n\textsuperscript{2}MFn\textsuperscript{2}, having input, output and a single hidden layer \cite{dutta2022neural, dutta2022n} has been developed for dimension reduction, based on the notion of NMF. Two deep neural network models, called Deep Neural Network for Non-negative Matrix Factorization (DN3MF) \cite{duttaPAAdn3mf} and Multiple Deconstruction Single Reconstruction Deep Neural Network Model for Non-negative Matrix Factorization (MDSR-NMF) \cite{prasun2023MDSR}, have also been developed with the aim of low rank approximation.

The aspiration of simulating the factorization behaviour of the traditional NMF technique ensuring the outcome of a unique pair of factor matrices of the reconstructed input matrix has motivated us for the progressive development of the previous three models. In this work, we have fused the advantages of conventional iterative learning with those of deep learning in a way that resembles the trait of human learning. While learning, humans always attempt to disintegrate the concepts into smaller fragments and try to learn hierarchically referring back to the original details frequently ensuring the correctness of the learning. Thus, we can claim that human learning is always input-guided.

In this paper, we have designed a model, called Input Guided Multiple Deconstruction Single Reconstruction neural network for Non-negative Matrix Factorization (IG-MDSR-NMF) towards dimension reduction. Deconstruction and reconstruction are the two phases of the model. The layers in the deconstruction phase receive the hierarchically processed output of the preceding layer along with a copy of the original data as input. Thus the model is called ``Input Guided". There is only one layer in the reconstruction phase, which ensures a true realization of the NMF technique generating a unique pair of non-negative factor matrices.

The main objective of the study is to find a low rank approximation of the input data to get rid of the curse of dimensionality problem. Constraints such as the non-negativity of both the factor matrices limit the learning of the model to some extent. Relaxing the non-negativity criteria of the coefficient matrix does not hamper the overall aim of the model, i.e., dimension reduction. The input as well as its low rank approximation, i.e., the basis matrix adheres to the non-negativity constraint, only the non-negativity restriction of the coefficient matrix is relaxed. This novel idea has been called ``Relaxed Non-negative Matrix Factorization (RNMF)". The model realizing the same has been named Input Guided Multiple Deconstruction Single Reconstruction neural network for Relaxed Non-negative Matrix Factorization (IG-MDSR-RNMF).

The quality of low dimensional embedding produced by IG-MDSR-NMF/ IG-MDSR-RNMF has been verified based on the extent of input shape preservation. The need for dimension reduction over the original data has also been judged and justified. The superiority of the low rank approximation by IG-MDSR-NMF/IG-MDSR-RNMF has also been verified over nine well-known dimension reduction techniques, experimenting on five datasets for both classification and clustering. The efficacy of IG-MDSR-NMF/IG-MDSR-RNMF has been justified for downstream analyses (classification and clustering) on different types of datasets.

The remaining sections of the paper are organised as follows. Section 2 discusses the design and learning of the proposed deep learning models, IG-MDSR-NMF and IG-MDSR-RNMF, including their computational complexity. Section 3 starts with summarising the datasets, followed by demonstrating the technicalities of the experimental setup and procedure, and presents the results with an appropriate analysis. Finally, it includes a convergence study for both IG-MDSR-NMF and IG-MDSR-RNMF. Subsequently, Section 4 takes the paper to its conclusion.

\section{IG-MDSR-NMF and IG-MDSR-RNMF}
\label{model_IG-MDSR-NMF}
This section describes the philosophy behind the novel architecture of the Input Guided Multiple Deconstruction Single Reconstruction neural network for Non-negative Matrix Factorization (IG-MDSR-NMF) model. This is followed by a design of its architecture and learning strategy. IG-MDSR-NMF is a true realization of NMF, where both the resulting factors, i.e., the basis matrix and the coefficient matrix follow the non-negativity criteria. A relaxed version of the model, called Input Guided Multiple Deconstruction Single Reconstruction neural network for Relaxed Non-negative Matrix Factorization (IG-MDSR-RNMF) has also been designed, where the non-negativity constraint of the coefficient matrix has been relaxed. That is, in IG-MDSR-RNMF, the basis matrix adheres to the non-negativity constraint, whereas the coefficient matrix is unconstrained. The motivation, design, and learning strategy of IG-MDSR-RNMF are also described here.

\subsection{Motivation}
\label{motivation}
The philosophy behind the novel architecture(s) of the Input Guided Multiple Deconstruction Single Reconstruction neural network for (Relaxed) Non-negative Matrix Factorization model is described here.

\subsubsection{IG-MDSR-NMF}
\label{motivation_IG-MDSR-NMF}
Learning of human beings is an iterative process where complex concepts are gradually broken down into simpler ones. Along with the fragmented concepts, humans always refer to the original material whenever needed, ensuring correctness of learning. A Deep neural network learns hierarchically, i.e., the representation of data is learned in a layerwise approach. That is, the layer $l_1$ connecting the input layer $l_0$ learns directly from the input and the next layer $l_2$ learns from the learned representation of $l_1$ and so on. Therefore, the deeper the network grows, and the upper-level layers receive a more abstract form of representation of the original input. If an intermediate layer, for some reason, is unable to learn appropriately from its previous layer, this improper learning will percolate to the succeeding layers and finally will affect decision making to a great extent. Thus, there is a possibility of deviation in learning the actual information content of the raw data. To overcome this phenomenon, a novel architecture has been designed in this article, where the input layer $l_0$ is additionally connected with the layers $l_2, l_3, ...,$ till the second last layer of the architecture. The design enforces the model to be guided by the original representation of the input in the process of hierarchically learning representation of the same, mimicking the human way of learning. As each layer of the model is guided by the input, the model has been named as Input Guided MDSR-NMF.

\textit{\\Realization of the concept\\\\}
IG-MDSR-NMF has been developed for the task of NMF. The model is divided into two phases, viz., deconstruction and reconstruction. The input to the neural network is transformed into latent space during the deconstruction phase, and the network attempts to reconstruct the input from its low rank representation during the reconstruction phase. There are multiple deconstruction layers leading towards the slenderest layer of the network from the input but there is only one reconstruction layer connecting the slenderest and the output layers of the network. Hence, the model is called Multiple Deconstruction Single Reconstruction neural network model. The novel architecture tries to learn the low rank representation of the input data in a stepwise manner in the deconstruction phase of the model. Whereas, in the reconstruction phase, it directly tries to reconstruct the input from the latent space. The architecture has been designed in this manner to simulate the factorization behaviour of the traditional NMF technique. By taking only one layer in the reconstruction step, the model can synthesize a unique pair of constituent non-negative factor matrices.

The model aims to factorise a given input matrix into two constituent factor matrices using Non-negative Matrix Factorization (NMF) aiming towards dimension reduction. The input matrix and both the factor matrices satisfy the non-negativity criterion. In IG-MDSR-NMF, the input matrix $\mathbf{X}$ is factored to produce two factor matrices namely, $\mathbf{B}$ and $\mathbf{W}$, where $\mathbf{B}$ is the output of the slenderest layer of the architecture, and $\mathbf{W}$ is the weight matrix connecting the slenderest layer and the output layer of the model. $\mathbf{B}$ is called the basis matrix and $\mathbf{W}$ represents the coefficient matrix. These two factors are used to compute $\mathbf{\hat{X}}$ as $\mathbf{\hat{X}}= \mathbf{BW}$, where $\mathbf{X}$ is regenerated by $\mathbf{\hat{X}}$. Thus, we need to produce a unique pair of $\mathbf{B}$ and $\mathbf{W}$ as the output of the model.

IG-MDSR-NMF attempts to learn the low rank representation of the input cumulatively, leveraging the benefits of hierarchical learning enabled by the deep neural network architecture. Hence, there is more than one layer in the deconstruction phase of the model and a unique $\mathbf{B}$ is available as the output at the end of the deconstruction phase. The model attempts to imitate the factorization behaviour of the traditional NMF technique by taking only one layer in the reconstruction phase of the model. If there were multiple layers in the reconstruction phase then finding a unique value of $\mathbf{W}$ with respect to $\mathbf{X}$ and $\mathbf{B}$ would not be possible. Let there be $k$ layers in the reconstruction phase. The weight matrices are denoted as $\mathbf{W}_1, \mathbf{W}_2,...,\mathbf{W}_k$ accordingly. For the sake of simplicity, we ignore the activation functions and write
\begin{equation}
\mathbf{\hat{X}}=\mathbf{B} \mathbf{W}_1 \mathbf{W}_2 ... \mathbf{W}_k
\label{ig_xhat}
\end{equation}
Now, from equation \eqref{ig_xhat} we need to obtain a unique $\mathbf{W}$, which can be defined as $\mathbf{W}=\prod_{i=1}^{k} \mathbf{W}_i$.
Thus, we can write
\begin{equation}
\mathbf{W}=\mathbf{B}^{-1} \mathbf{\hat{X}}
\label{ig_w}
\end{equation}
The input matrix $\mathbf{X}$ is not always a square matrix. Thus, $\mathbf{\hat{X}}$ and $\mathbf{B}$ are not also square in nature. In this situation, we can compute the pseudoinverse, which is not unique. Thus, multiple reconstruction layers fail to compute a unique $\mathbf{W}$, whereas a single reconstruction layer ensures a unique pair of $\mathbf{B}$ and $\mathbf{W}$ as the output. The low rank representation of $\mathbf{X}$ is $\mathbf{B}$.

\subsubsection{IG-MDSR-RNMF}
\label{motivation_IG-MDSR-RNMF}
Our main objective is to find a low dimensional non-negative representation of the input matrix $\mathbf{X}$. The low rank representation $\mathbf{B}$ should embed relevant information in a computationally efficient manner. The second factor $\mathbf{W}$ is only used to regenerate the input $\mathbf{X}$. In IG-MDSR-RNMF we have relaxed the non-negativity constraint of the coefficient matrix $\mathbf{W}$, whereas the non-negativity constraint of the basis matrix $\mathbf{B}$ remains intact. Finding a set of non-negative factor matrices is like confronting the estimation of the low dimensional embedding by enforcing the non-negativity criteria of the coefficient matrix. The relaxation does not compromise with the ultimate objective of finding the non-negative low dimensional representation of the input. Whereas, relaxation helps manoeuvre the learning more efficiently to find the best possible low rank representation of the input. Relaxing the non-negativity criteria of one of the factor matrices has been presented as a new class of matrix factorization technique called, Relaxed Non-negative Matrix Factorization (RNMF).

\subsection{Architecture}
\label{architecture_IG-MDSR-NMF}
IG-MDSR-NMF is a deep neural network architecture made up of an input layer, $s$ hidden layers and an output layer. Consider a given data matrix, $\mathbf{U} = [u_{pi}]_{m \times n'}$. We process $\mathbf{U}$ to generate a matrix $\mathbf{X} = [x_{pi}]_{m \times n}$ with each element being non-negative. We use a methodology similar to the folding data method described in \cite{kim2003subsystem} to carry out this task. This approach uses two columns of $\mathbf{X}$ to represent each column of $\mathbf{U}$. That is, $i^{th}$ and $(n'+i)^{th}$ columns of $\mathbf{X}$ correspond to $i^{th}$ column of $\mathbf{U}$. Entries in every column of $\mathbf{U}$ can be either positive or negative. Positive values from $i^{th}$ column of $\mathbf{U}$ are kept in $i^{th}$ column of $\mathbf{X}$, whereas the absolute form of the negative values is stored in $(n'+i)^{th}$ column of $\mathbf{X}$. The remaining empty cells in $i^{th}$ and $(n'+i)^{th}$ columns of $\mathbf{X}$ are filled with zeros. To obtain the original elements of the $i^{th}$ column of $\mathbf{U}$, subtract the elements of the $(n'+i)^{th}$ column of $\mathbf{X}$ from the elements of its $i^{th}$ column. As a result, the number of columns in $\mathbf{X}$ is exactly twice that of $\mathbf{U}$, i.e., $n = 2n'$. Furthermore, it should be noted that in this manner, exactly half of the elements of $\mathbf{X}$ are zero, resulting in a sparse matrix. Each row of $\mathbf{X}$ is now used as input to the model.

\begin{figure}[ht!]
\centering
\includegraphics[width=\linewidth\iffalse125mm\fi, scale=1.0]{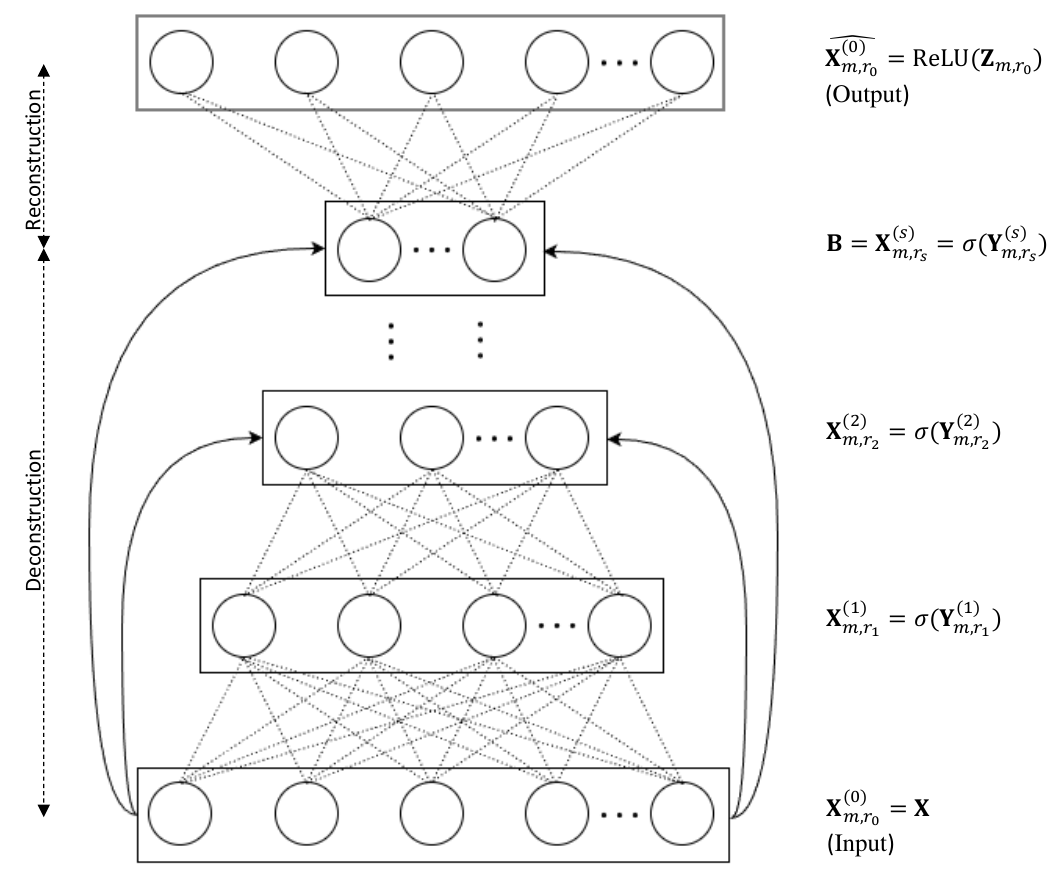}
\caption{The architecture of IG-MDSR-NMF}
\label{IG-MDSR-NMF_ProposedModel}
\end{figure}

The uppermost/last hidden layer of IG-MDSR-NMF has been envisioned to be the model's slender layer, with $r$ nodes that extract $r<n'$ features. The IG-MDSR-NMF design puts no restrictions on the value of the low rank dimension $r$ in relation to the number of samples $m$, i.e., $r<min(m,n')$, or $m<r<n'$. In contrast, PCA requires $r$ to be smaller than both $m$ and $n'$, i.e., $r<min(m, n')$ \cite{pedregosa2011scikit}. Semi-NMF and DS-NMF approaches restrict the value of $r$ to $2*r<min(m,n')$ \cite{trigeorgis2014deep, trigeorgis2016deep}. The responsibility of the output layer is to reconstruct the original data using the extracted features. The model may be separated into two phases - deconstruction and reconstruction. The input and $s$ hidden layers comprise the deconstruction phase. The reconstruction phase consists of the slenderest and output layers. The latent representation of the input is obtained at the end of the deconstruction step. During the reconstruction process, the model tries to regenerate the input from this latent representation. As there is more than one deconstruction layer, but only one reconstruction layer, the design is referred to as multiple deconstruction, single reconstruction deep learning architecture. Figure \ref{IG-MDSR-NMF_ProposedModel} depicts the architecture of IG-MDSR-NMF.

In IG-MDSR-NMF, we have employed three different types of activation functions. After preprocessing, data is put into the network. To follow the typical neural network architecture, an identity function has been used as the activation function for the input nodes. Thus, the output of any input node is the same as its input. The sigmoid function has been employed for the hidden layers, whereas the ReLU activation function has been used for the output layer nodes. Sigmoid function maps the input of any interval $(-\infty, +\infty)$ to $(0, 1)$ as output. In contrast, all negative values are discarded with zero by the ReLU activation function. Since ReLU eliminates all negative components in order to meet non-negativity, the ReLU activation function experiences data loss, while the sigmoid function does not. Therefore, the model's non-negativity condition has been met by using the sigmoid and ReLU activation functions, and the data loss issue has been avoided by using the sigmoid function in the hidden layers.

We consider the input layer as the $0^{th}$ layer of the model having $r_0=n$ number of nodes. The input to this layer is denoted as $\mathbf{X}^{(0)}=[x^{(0)}_{pi_0}]_{m \times r_0}$, where $\mathbf{X}^{(0)} = \mathbf{X}$. The input layer connects the first hidden layer of the model having $r_1<r_0$ number of nodes. The output of the first hidden layer is now expressed in matrix form for all samples as $\mathbf{X}^{(1)}=[x^{(1)}_{pi_1}]_{m \times r_1}$ and is defined as
\begin{equation}
\mathbf{X}^{(1)} = \sigma(\mathbf{Y}^{(1)})
\label{ig_x1}
\end{equation}
where, $\sigma(\mathbf{Y}^{(1)})$ is an $m \times r_1$ matrix and each element of the matrix is computed by applying the sigmoid activation function ($\sigma$) to the corresponding element of $\mathbf{Y}^{(1)}$, where $\mathbf{Y}^{(1)} = [y^{(1)}_{pi_1}]_{m \times r_1}$ is defined as
\begin{equation}
\mathbf{Y}^{(1)} = \mathbf{X}^{(0)} \mathbf{V}^{(1)}
\label{ig_y1}
\end{equation}
Here $\mathbf{V}^{(1)} = [v^{(1)}_{i_0i_1}]_{r_0 \times r_1}$ is the weight matrix between the input and the first hidden layers. The output of the first hidden layer connects to the second hidden layer containing $r_2$ nodes, where $r_2<r_1$. Additionally, the input layer is also connected to this second hidden layer. The weight matrix between the first and second hidden layers is denoted by $\mathbf{V}^{(2)} = [v^{(2)}_{i_1i_2}]_{r_1 \times r_2}$ and the weight matrix between the input and second hidden layers is $\Tilde{\mathbf{V}}^{(2)} = [\Tilde{v}^{(2)}_{i_0i_2}]_{r_0 \times r_2}$. The output of the second hidden layer is expressed as $\mathbf{X}^{(2)}=[x^{(2)}_{pi_2}]_{m \times r_2}$ and is defined as
\begin{equation}
\mathbf{X}^{(2)} = \sigma(\mathbf{Y}^{(2)})
\label{ig_x2}
\end{equation}
where, $\mathbf{Y}^{(2)} = [y^{(2)}_{pi_2}]_{m \times r_2}$ is computed as
\begin{equation}
\mathbf{Y}^{(2)} = \mathbf{X}^{(1)} \mathbf{V}^{(2)} + \mathbf{X}^{(0)} \Tilde{\mathbf{V}}^{(2)}
\label{ig_y2}
\end{equation}
Similarly, the third hidden layer is connected to the output of the second hidden layer and the input layer. Eventually, the slenderest layer, i.e., the $s^{th}$ hidden layer of the model is connected to the output of the $(s-1)^{th}$ hidden layer and the input layer. The number of nodes in this slenderest layer is $r_s=r$. It is to be noted that $r=r_s<r_{s-1}<...<r_2<r_1<r_0=n$. The weight matrix between the $s^{th}$ and $(s-1)^{th}$ hidden layers is denoted by $\mathbf{V}^{(s)} = [v^{(s)}_{i_{s-1}i_s}]_{r_{s-1} \times r_s}$ and the weight matrix between the input and $s^{th}$ hidden layers is $\Tilde{\mathbf{V}}^{(s)} = [\Tilde{v}^{(s)}_{i_0i_s}]_{r_0 \times r_s}$. The output of this slenderest layer is denoted by $\mathbf{B}=\mathbf{X}^{(s)}$, where $\mathbf{X}^{(s)}=[x^{(s)}_{pi_s}]_{m \times r_s}$ is defined as
\begin{equation}
\mathbf{X}^{(s)} = \sigma(\mathbf{Y}^{(s)})
\label{ig_xs}
\end{equation}
Here, $\mathbf{Y}^{(s)} = [y^{(s)}_{pi_s}]_{m \times r_s}$ is determined as
\begin{equation}
\mathbf{Y}^{(s)} = \mathbf{X}^{(s-1)} \mathbf{V}^{(s)} + \mathbf{X}^{(0)} \Tilde{\mathbf{V}}^{(s)}
\label{ig_ys}
\end{equation}
The slenderest layer concludes the deconstruction phase of the model and marks the beginning of its reconstruction phase. The output of the slenderest layer, $\mathbf{B}$, represents the low rank representation of the input $\mathbf{X}$. The reconstruction phase comprises a single reconstruction layer, producing the output of the model, i.e., the regenerated input $\widehat{\mathbf{X}}=\widehat{\mathbf{X}^{(0)}}$. Here, $\widehat{\mathbf{X}^{(0)}}=[\widehat{x^{(0)}_{pi_0}}]_{m \times r_0}$ is computed as
\begin{equation}
\widehat{\mathbf{X}^{(0)}} = ReLU(\mathbf{Z})
\label{ig_x0hat}
\end{equation}
where we get $\mathbf{Z} = [z_{pi_0}]_{m \times r_0}$ as
\begin{equation}
\mathbf{Z} = \mathbf{X}^{(s)} \mathbf{W}
\label{ig_z}
\end{equation}
Here, $\mathbf{W} = [w_{i_si_0}]_{r_s \times r_0}$ represents the weight matrix between the slenderest layer and the output layer of the model.

The elements of the weight matrices $\mathbf{V}^{(1)}$, $\mathbf{V}^{(2)}$, ..., $\mathbf{V}^{(s)}$ and $\Tilde{\mathbf{V}^{(2)}}$, $\Tilde{\mathbf{V}^{(3)}}$, ..., $\Tilde{\mathbf{V}^{(s)}}$ are unrestricted, while the elements of the weight matrix $\mathbf{W}$ must be non-negative to meet the non-negativity requirement of the NMF algorithm. The two non-negative components of the regenerated input matrix $\mathbf{\widehat{X}}$ are the slender layer output $\mathbf{B}$ and the weight matrix $\mathbf{W}$.

The architecture of IG-MDSR-RNMF is the same as that of IG-MDSR-NMF, only the non-negativity requirement of the weight matrix $\mathbf{W}$ has been relaxed. That is the slender layer output $\mathbf{B}$ will follow the non-negativity requirement but, the weight matrix $\mathbf{W}$ connecting the slender layer and the output layer will be unconstrained.

\subsection{Learning}
\label{learning_IG-MDSR-NMF}
The objective of both IG-MDSR-NMF and IG-MDSR-RNMF is to find the best possible reconstruction ($\hat{\mathbf{X}}$) of the input matrix ($\mathbf{X}$) while factorizing $\mathbf{X}$ into $\mathbf{B}$ and $\mathbf{W}$. Thus, the objective function is defined as the mean square loss of the original and the regenerated input, i.e., we want to minimize $\|\mathbf{X} - \mathbf{\widehat{X}}\|_F$. Thus, the cost function $\Phi$ is defined as
\begin{equation}
\Phi = \frac{1}{2mn} \sum_{p=1}^m \sum_{j=1}^n (x_{pj} - \widehat{x}_{pj})^2
\label{ig_phi}
\end{equation}
IG-MDSR-NMF and IG-MDSR-RNMF have been trained using a stochastic gradient descent method based on adaptive estimation of first-order and second-order moments employing the Adam optimisation technique \cite{kingma2014adam}. The use of sigmoid activation function in the hidden layers of the network ensures the non-negativity requirement of the latent space representation ($\mathbf{B}$) of the input data matrix. Similarly, the ReLU activation function in the output layer guarantees the non-negativity of the regenerated input matrix $\hat{\mathbf{X}}$. In IG-MDSR-NMF, the non-negativity of the other factor matrix, i.e., the weight matrix $\mathbf{W}$ has been assured by replacing the negative elements arising in the course of the updation of weights during backpropagation with zeros.

\subsection{Computational Complexity}
\label{complexity_IG-MDSR-NMF}
The computational complexity of IG-MDSR-NMF and IG-MDSR-RNMF has been measured in terms of the number of operations performed. Both IG-MDSR-NMF and IG-MDSR-RNMF have $s+1$ deconstruction layers and one reconstruction layer. The input, i.e., each row of $\mathbf{X}$, travels through the identity function (activation function) at the input layer, hence the computational complexity is $\mathcal{O}(mr_0)$. The following step is defined in equation~(\ref{ig_y1}), where the complexity is $\mathcal{O}(mr_0r_1)$. The value of $\mathbf{Y}^{(1)}$ is now sent via the activation function $\sigma$ (equation~\ref{ig_x1}), with the complexity of this operation being $\mathcal{O}(mr_1)$. The following step using equation~(\ref{ig_y2}) includes $\mathcal{O}(mr_1r_2+mr_0r_2)$ operations. Thereafter, similar to the previous layer the value of $\mathbf{Y}^{(2)}$ passes through the activation function $\sigma$ (equation~\ref{ig_x2}) contributing $\mathcal{O}(mr_2)$ to the overall complexity. Proceeding this way the model finally computes $\widehat{\mathbf{X}^{(0)}}$ (equations~\ref{ig_x0hat} and \ref{ig_z}), and for this the computational complexity is $\mathcal{O}(mr_sr_0+mr_0)$. Thus, the forward pass comprises $\mathcal{O}(mr_0+mr_0r_1+mr_1+mr1r_2+mr_0r_2+...+mr_sr_0+mr_0)$ operations. As mentioned before, $n=r_0 > r_1 > r_2 > ... > r_s=r$. Removing the lower order terms, the computational cost of the forward pass is $\mathcal{O}(mr_0r_1)$. The computational complexity of $\Phi$ (equation~\ref{ig_phi}) is $\mathcal{O}(mr_0)$. The major task of backward propagation is to update the weights. With similar arguments, we can conclude that the backward pass entails $\mathcal{O}(mr_0r_1)$ operations. Thus, the computational cost of an epoch is $\mathcal{O}(mr_0r_1)$. The complexity for $t$ such epochs is $\mathcal{O}(tmr_0r_1)$. Hence, the overall computational complexity of IG-MDSR-NMF and IG-MDSR-RNMF is $\mathcal{O}(tmnr_1)$.

\section{Experiments, Results and Analysis}
\label{experiments_IG-MDSR-NMF}
The efficacy of dimensionality reduction with IG-MDSR-NMF has been studied in two ways. First, the extent to which IG-MDSR-NMF can preserve the local structure of data after dimension reduction has been compared with that of nine different dimension reduction approaches. The quality of the low rank representation has also been assessed against the original dataset. Second, the quality of dimensionality reduction using IG-MDSR-NMF has been compared with that of nine different dimension reduction techniques in terms of classification and clustering. We have used five popular datasets for dimension reduction. Three of the nine dimension reduction methods are traditional dimension reduction methodologies; one is a classic NMF technique and the other five are current state-of-the-art neural network based NMF implementation techniques. The same testing procedure has been followed for IG-MDSR-RNMF to determine the effectiveness of the model.

Data sources are narrated in Section \ref{data_IG-MDSR-NMF}. Section \ref{exp_setup_IG-MDSR-NMF} provides the experimental setup for IG-MDSR-NMF and IG-MDSR-RNMF, as well as the weight initialization technique. The methodology to test the extent of local structure preservation is described in Section \ref{exp_procedure_IG-MDSR-NMF}. The experimental procedure for evaluating the quality of dimensionality reduction in terms of classification and clustering has been also described in Section \ref{exp_procedure_IG-MDSR-NMF}. Section \ref{analysis_IG-MDSR-NMF} assesses the comparative performance of IG-MDSR-NMF and IG-MDSR-RNMF over other dimension reduction methods. Section \ref{convergence_IG-MDSR-NMF} depicts a graphical representation of the convergence of the objective function for both IG-MDSR-NMF and IG-MDSR-RNMF.

\subsection{Data Sources}
\label{data_IG-MDSR-NMF}
Four popular datasets, viz., Gastrointestinal Lesions in Regular Colonoscopy (GLRC) dataset \cite{mesejo2016computer}, Online News Popularity (ONP) dataset \cite{fernandes2015proactive}, Parkinson's Disease Classification (PDC) dataset \cite{sakar2019comparative} and Student Performance (SP) dataset \cite{cortez2008using} have been downloaded from the UCI machine learning repository \cite{Dua:2019}. The MovieLens dataset has been acquired from the GroupLens research lab website. GroupLens is a research lab at the Department of Computer Science and Engineering at the University of Minnesota. The effectiveness of dimensionality reduction by IG-MDSR-NMF and IG-MDSR-RNMF have been evaluated using these datasets.

\subsubsection{Gastrointestinal Lesions in Regular Colonoscopy (GLRC) dataset}
The dataset \cite{mesejo2016computer} consists of ground truth and features extracted from a colonoscopic video database of gastrointestinal lesions. Expert image inspection and histology have been used for modelling the ground truth. The dataset has 76 samples. There are 698 features in all, with 2D textural properties of the lesions accounting for the first 422 attributes, 2D colour features of the lesions accounting for the next 76 and 3D shape features of the lesions accounting for the last 200. Each sample used two distinct types of lighting. We have considered the kind of light as a feature while performing the computation. Thus, the data matrix has $2 \times 76=152$ rows and $1+698=699$ columns. The dataset consists of three types of lesions: hyperplasic, adenoma and serrated adenoma. Hyperplastic lesions are benign, while, adenoma and serrated adenoma lesions are malignant. As a result, we treat the dataset as a two-class problem \cite{jiang2018direct}.

\subsubsection{Online News Popularity (ONP) dataset}
The dataset \cite{fernandes2015proactive} contains multiple sets of features extracted from Mashable articles published between January 7, 2013 and January 7, 2015. The collection includes 39644 entries for online articles. Each item is defined by a total of 60 features. URL is one of them, which we have not added because of its uniqueness to each article. The other 59 features are numeric, based on the article's structure and content, with the last one being the number of days between article publication and dataset acquisition. As a consequence, the dataset size is $39644 \times 59$. The samples in the dataset are divided into two categories namely, popular and unpopular, based on the number of shares of each article \cite{laugel2017inverse, laugel2018comparison}.

\subsubsection{Parkinson's Disease Classification (PDC) dataset}
A study employing voice recordings of 252 individuals was undertaken at the Department of Neurology at Cerrahpaşa Faculty of Medicine, Istanbul University, Istanbul, Turkey \cite{sakar2019comparative}. There were 188 patients with Parkinson's disease, while the remaining 64 did not have the ailment. As a consequence, the samples are divided into two groups. Each sample is described by 754 features. Time-frequency features, mel-frequency cepstral coefficients, wavelet transform-based features, vocal fold features and wavelet transform features with a configurable Q factor were employed. These characteristics have provided clinically relevant information for Parkinson's disease assessment. We have not considered the patient identification number as a feature. The experiment was repeated three times for each individual. Thus, the data matrix size is $756 \times 753$.

\subsubsection{Student Performance (SP) dataset}
The dataset \cite{cortez2008using} includes student data collected from two Portuguese secondary schools in the Alentejo region of Portugal in 2005 and 2006. It uses data from two sources: student grade reports and student replies to a series of questionnaires. The features include different student grades, as well as demographic, social and school-related information. Here, we have considered the Math performance of 395 students. These data are represented by 29 different features. The dataset has a feature identifying the student's school. Each student receives three unique grades: G1, G2 and G3 representing the first period grade, second period grade and final grade respectively. Grades are simply numerical values ranging from 0 to 20. G1 and G2 have been viewed as features, whereas G3 has been used as a target attribute. Thus, each student has been represented by 32 features, resulting in a database size of $395 \times 32$. We have designed the problem as a two-class problem based on G3, with a student passing if G3 is more than or equal to 10 and failing otherwise.

\subsubsection{MovieLens dataset}
The MovieLens datasets were prepared by the members of the GroupLens Research Project at the University of Minnesota \cite{harper2015movielens}. This dataset comprises 100,000 ratings for 1682 movies from 943 people and the remaining entries are unavailable. Thus, the dataset comprises $943$ samples, each having $1682$ features. The ratings range from 1 to 5, with 1 being the lowest rating and 5 being the highest. Every user has rated at least twenty movies. The dataset also includes basic demographic information such as the users' age, gender, employment and zip code. The data was gathered over seven months, from September 19, 1997, to April 22, 1998, using the MovieLens website (movielens.umn.edu). Users with less than 20 ratings or with missing demographic information were removed from this dataset during the cleaning process. We have considered gender as the classifying attribute, hence the dataset is classified as a two-class problem.

\subsection{Experimental setup}
\label{exp_setup_IG-MDSR-NMF}
IG-MDSR-NMF and IG-MDSR-RNMF have been implemented in Keras (version 2.13.1). The Xavier normal initialization approach \cite{glorot2010understanding} is an effective weight initialization technique for neural networks with sigmoid activation functions. The elements of all weight matrices in the proposed IG-MDSR-NMF/IG-MDSR-RNMF model have been initialised using the same. Different software libraries, such as Scikit-learn, TensorFlow and Keras have been employed as and when needed to program several other existing dimensionality reduction techniques. Plots of various figures have been generated using Python programming language.

The data matrices have been preprocessed before using them as input to the models. Data matrices have been normalised using the Z score normalisation technique. If a classification/clustering performance score generates an error, i.e., fails to produce a valid output for any reason, the lowest possible value of that metric is assigned in that place during computation. This step ensures that the classification/clustering performance scores for various dimension reduction algorithms for a given dataset are consistent.

IG-MDSR-NMF/IG-MDSR-RNMF model has the input layer, $s$ hidden layers and the output layer. In our implementation, we have considered $s=3$. The number of training epochs is determined dynamically; training ends when the difference in the cost values in two consecutive epochs reaches a predetermined threshold.

\subsection{Experimental Procedure}
\label{exp_procedure_IG-MDSR-NMF}
Nine state-of-the-art dimension reduction techniques have been used to compare the performance of IG-MDSR-NMF. They include Autoencoder (AE) (with one hidden layer, the number of nodes in the hidden layer being the dimension of the transformed space), Principal Component Analysis (PCA), Uniform Manifold Approximation and Projection (UMAP), traditional NMF, Multiple Deconstruction Single Reconstruction Deep Neural Network Model for Non-negative Matrix Factorization (MDSR-NMF) \cite{prasun2023MDSR}, Deep Neural Network for Non-negative Matrix Factorization (DN3MF) \cite{duttaPAAdn3mf}, Non-negative Matrix Factorization Neural Network (n\textsuperscript{2}MFn\textsuperscript{2}) \cite{dutta2022neural, dutta2022n}, Semi-NMF \cite{trigeorgis2014deep} and Deep Semi-NMF (DS-NMF) \cite{trigeorgis2016deep}. Let these $10$ dimensionality reduction techniques, including IG-MDSR-NMF, be placed in set $T$ and a dimension reduction technique (i.e., an element of $T$) be denoted by $\dot{T}$. The IG-MDSR-RNMF model has also been subjected to a similar process.

A dataset $\mathbf{X}$, having $m$ samples and $n'$ features per sample, is reduced to a dimension $r$ ($r<n'$) determined using a random factor $f\;(0<f<1)$ and computed as $r = \lfloor n' \times f \rfloor$. For a certain value of $f$, the dataset $\mathbf{X}$ is dimensionally reduced using all $\dot{T}$ in $T$. Therefore, there are ten dimensionally transformed datasets for a factor $f$, each having the same $r$ value, denoted by $\mathbf{X}_r(\dot{T})$. The effectiveness of IG-MDSR-NMF will now be illustrated on these transformed datasets. The same procedure has also been followed for the IG-MDSR-RNMF algorithm.

The effectiveness of dimension reduction by IG-MDSR-NMF/IG-MDSR-RNMF has been demonstrated by measuring the quality of local structure retention and downstream analyses using classification and clustering. The quality of preservation of the local structure of data in low rank distribution by IG-MDSR-NMF/IG-MDSR-RNMF over that of other dimension reduction techniques has been judged using the trustworthiness score.

\begin{figure}[ht!]
\centering
\includegraphics[width=1.0\textwidth, scale=1.0]{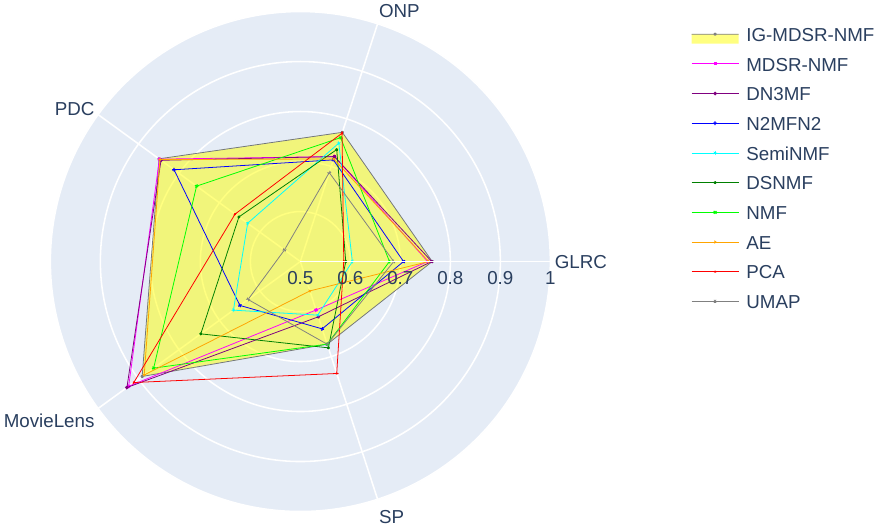}
\caption{Trustworthiness scores of ten dimension reduction techniques including IG-MDSR-NMF.}
\label{IG-MDSR-NMF_trustworthyness_V31}
\end{figure}
\begin{figure}[ht!]
\centering
\includegraphics[width=1.0\textwidth, scale=1.0]{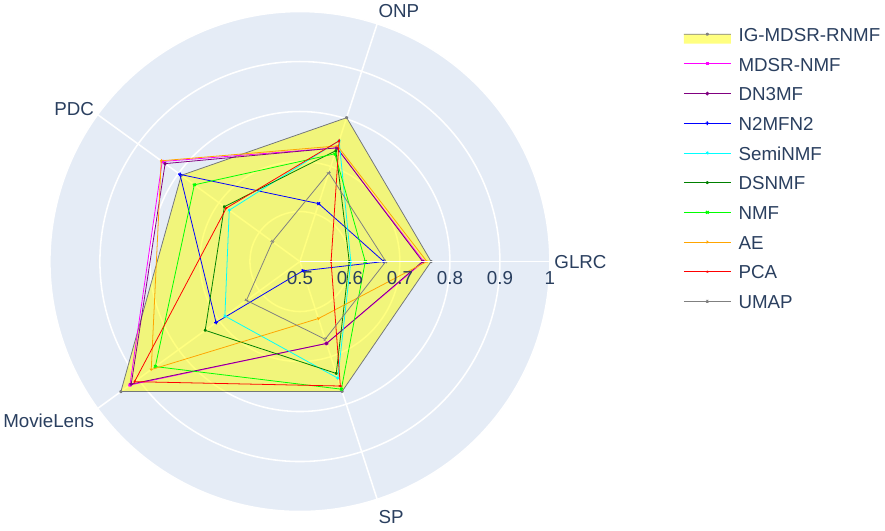}
\caption{Trustworthiness scores of ten dimension reduction techniques including IG-MDSR-RNMF.}
\label{IG-MDSR-RNMF_trustworthyness_V33}
\end{figure}

For downstream analyses, each reduced dataset $\mathbf{X}_r(\dot{T})$ has been classified using four well-known classification methods: K-Nearest Neighbor (KNN), Multilayer Perceptron (MLP), Naive Bayes (NB) and Quadratic Discriminant Analysis (QDA). Four classification performance metrics have been utilized to examine the quality of the classification done by the aforementioned classifiers: Accuracy (ACC), Cohen-Kappa score (CKS), F1 score (FS) and Matthews Correlation Coefficient (MCC). Thus, for each $\mathbf{X}_r(\dot{T})$, we get a classification performance score by performing classification using a classifier and validating the outcome using a classification performance metric. The same procedure has been carried out for the IG-MDSR-RNMF algorithm separately. The performance of IG-MDSR-NMF and IG-MDSR-RNMF has also been compared with that of the original dataset to demonstrate the superiority of the dimensionally reduced dataset over the original data, establishing the necessity of dimension reduction.

\begin{figure}[ht!]
\centering
\includegraphics[width=\linewidth\iffalse125mm\fi, scale=1.0]{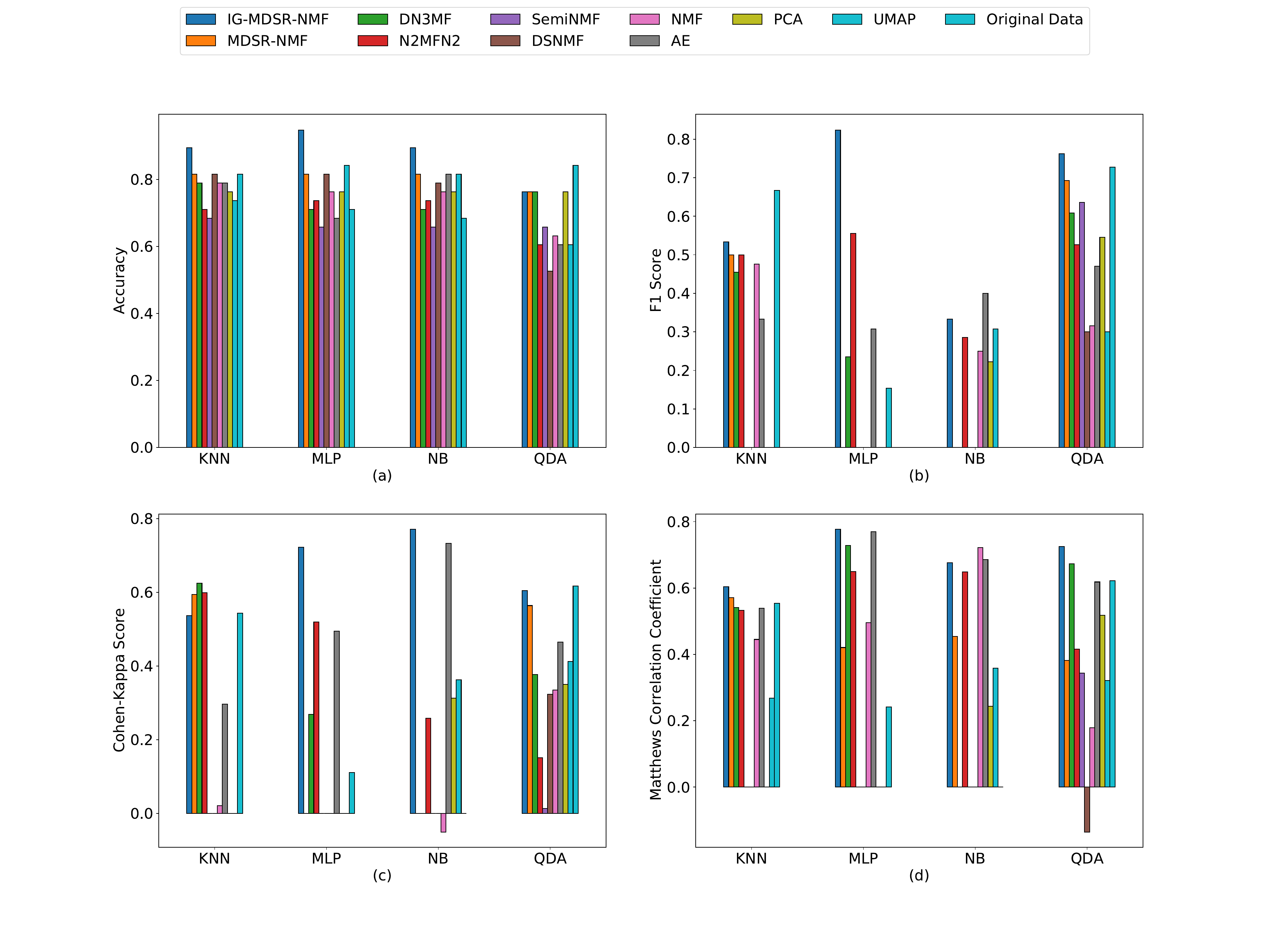}
\caption{Performance scores of the classification algorithms on the dimensionally reduced dataset instances of the GLRC dataset by IG-MDSR-NMF and nine other dimension reduction techniques along with the original data.}
\label{IG-MDSR-NMF_classification_GLRC_V31}
\end{figure}
\begin{figure}[ht!]
\centering
\includegraphics[width=\linewidth\iffalse125mm\fi, scale=1.0]{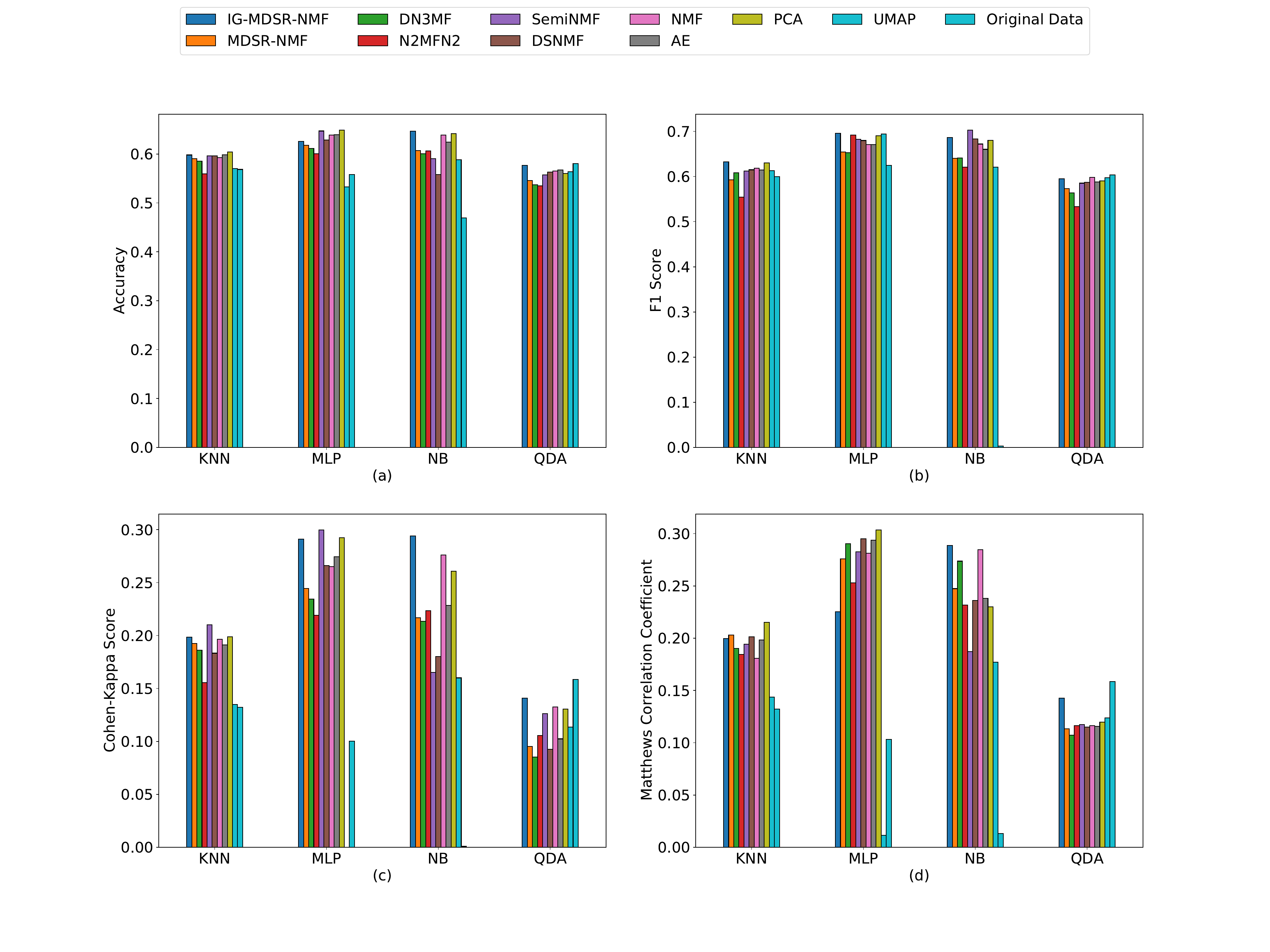}
\caption{Performance scores of the classification algorithms on the dimensionally reduced dataset instances of the ONP dataset by IG-MDSR-NMF and nine other dimension reduction techniques along with the original data.}
\label{IG-MDSR-NMF_classification_ONP_V31}
\end{figure}
\begin{figure}[ht!]
\centering
\includegraphics[width=\linewidth\iffalse125mm\fi, scale=1.0]{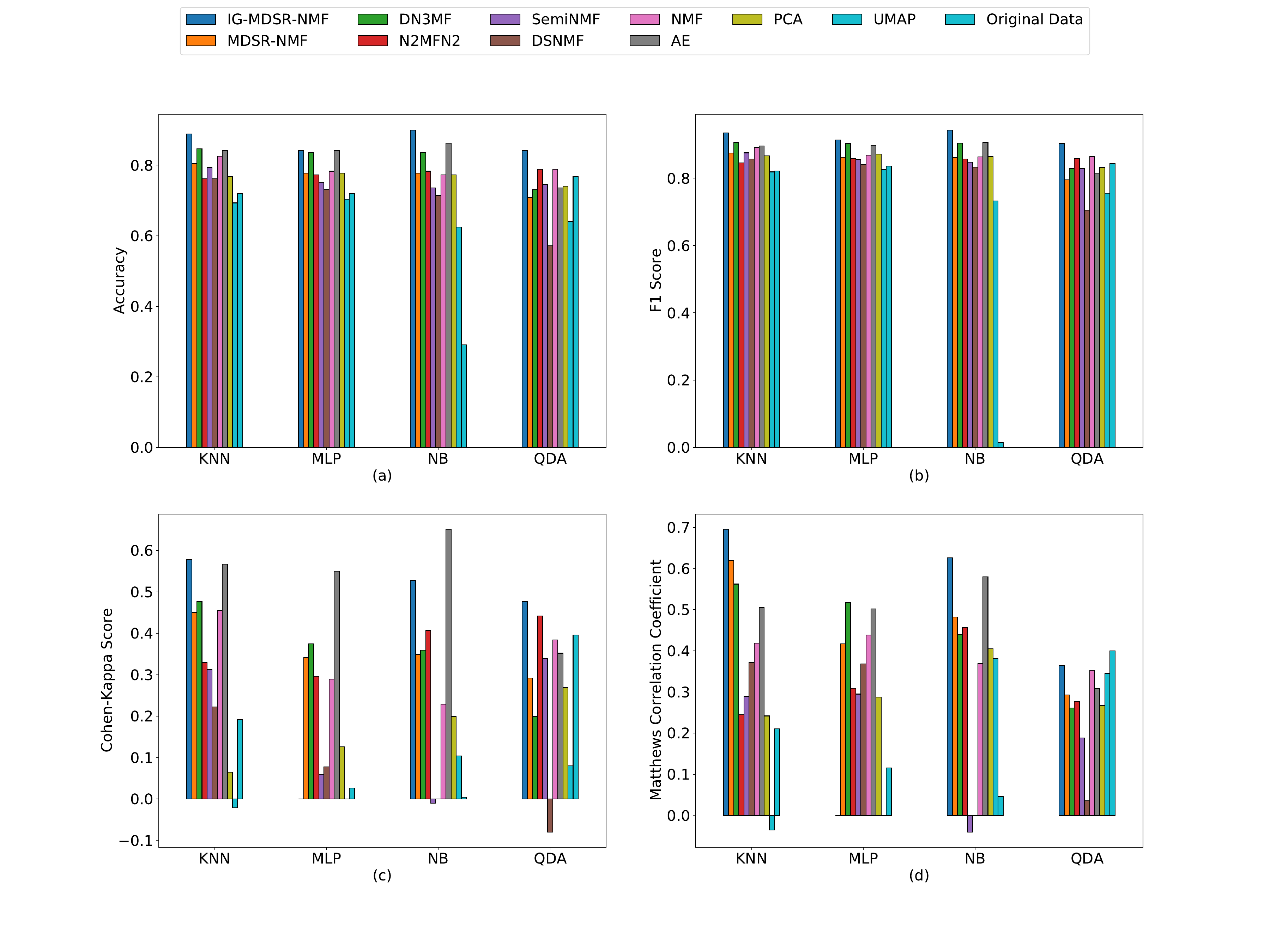}
\caption{Performance scores of the classification algorithms on the dimensionally reduced dataset instances of the PDC dataset by IG-MDSR-NMF and nine other dimension reduction techniques along with the original data.}
\label{IG-MDSR-NMF_classification_PDC_V31}
\end{figure}
\begin{figure}[ht!]
\centering
\includegraphics[width=\linewidth\iffalse125mm\fi, scale=1.0]{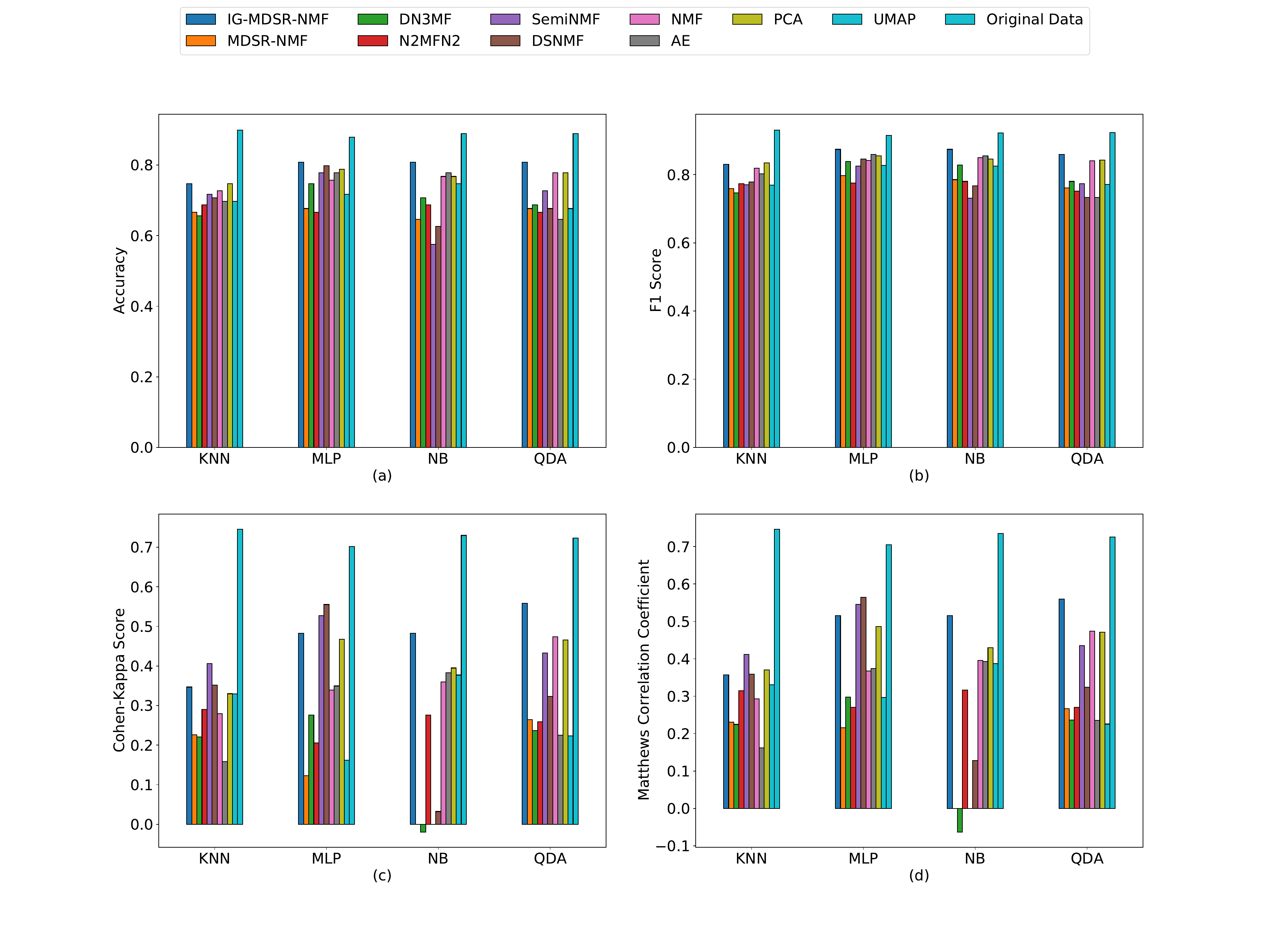}
\caption{Performance scores of the classification algorithms on the dimensionally reduced dataset instances of the SP dataset by IG-MDSR-NMF and nine other dimension reduction techniques along with the original data.}
\label{IG-MDSR-NMF_classification_SP_V31}
\end{figure}
\begin{figure}[ht!]
\centering
\includegraphics[width=\linewidth\iffalse125mm\fi, scale=1.0]{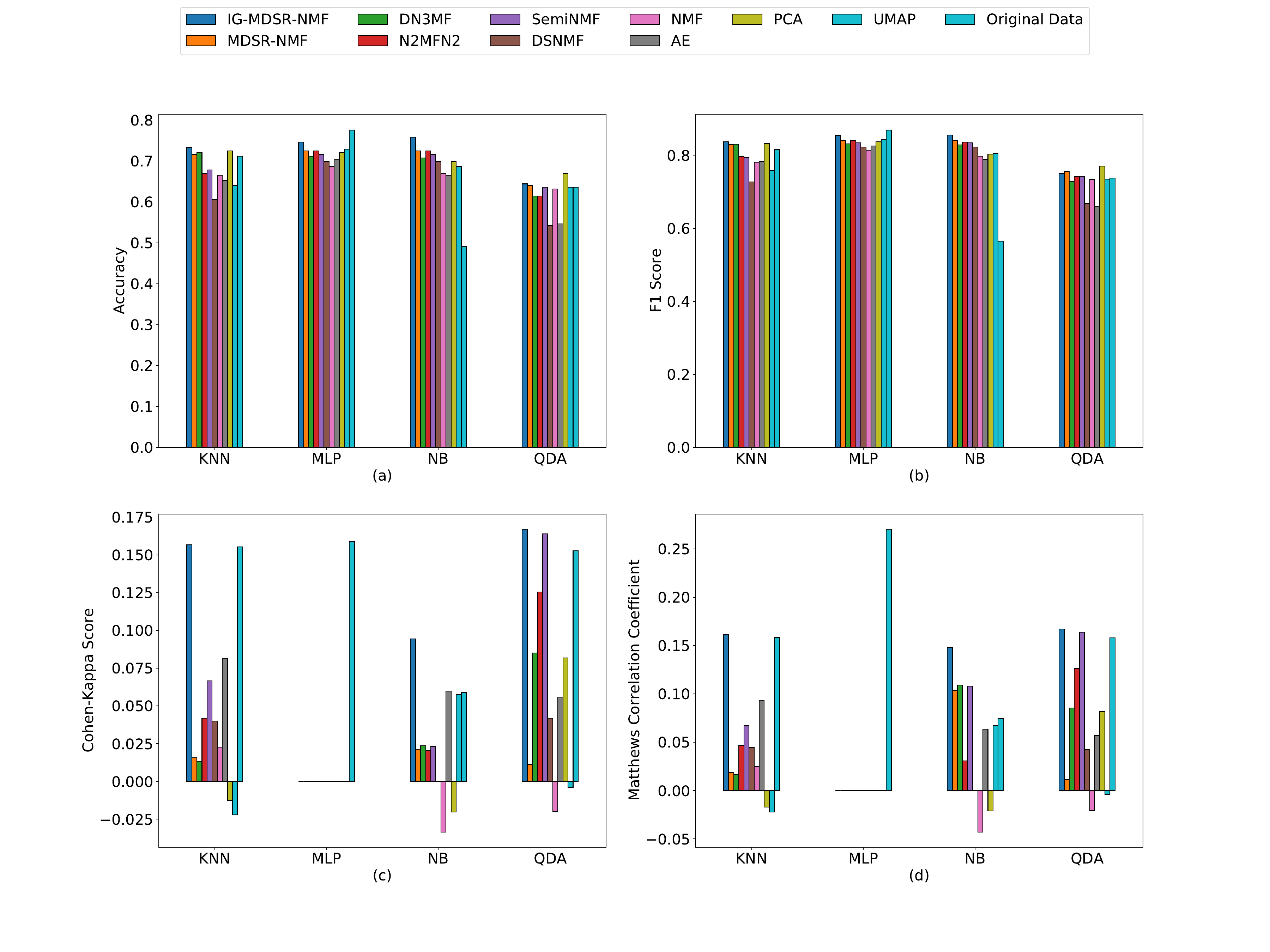}
\caption{Performance scores of the classification algorithms on the dimensionally reduced dataset instances of the MovieLens dataset by IG-MDSR-NMF and nine other dimension reduction techniques along with the original data.}
\label{IG-MDSR-NMF_classification_ml-100k_V31}
\end{figure}

\begin{figure}[ht!]
\centering
\includegraphics[width=\linewidth\iffalse125mm\fi, scale=1.0]{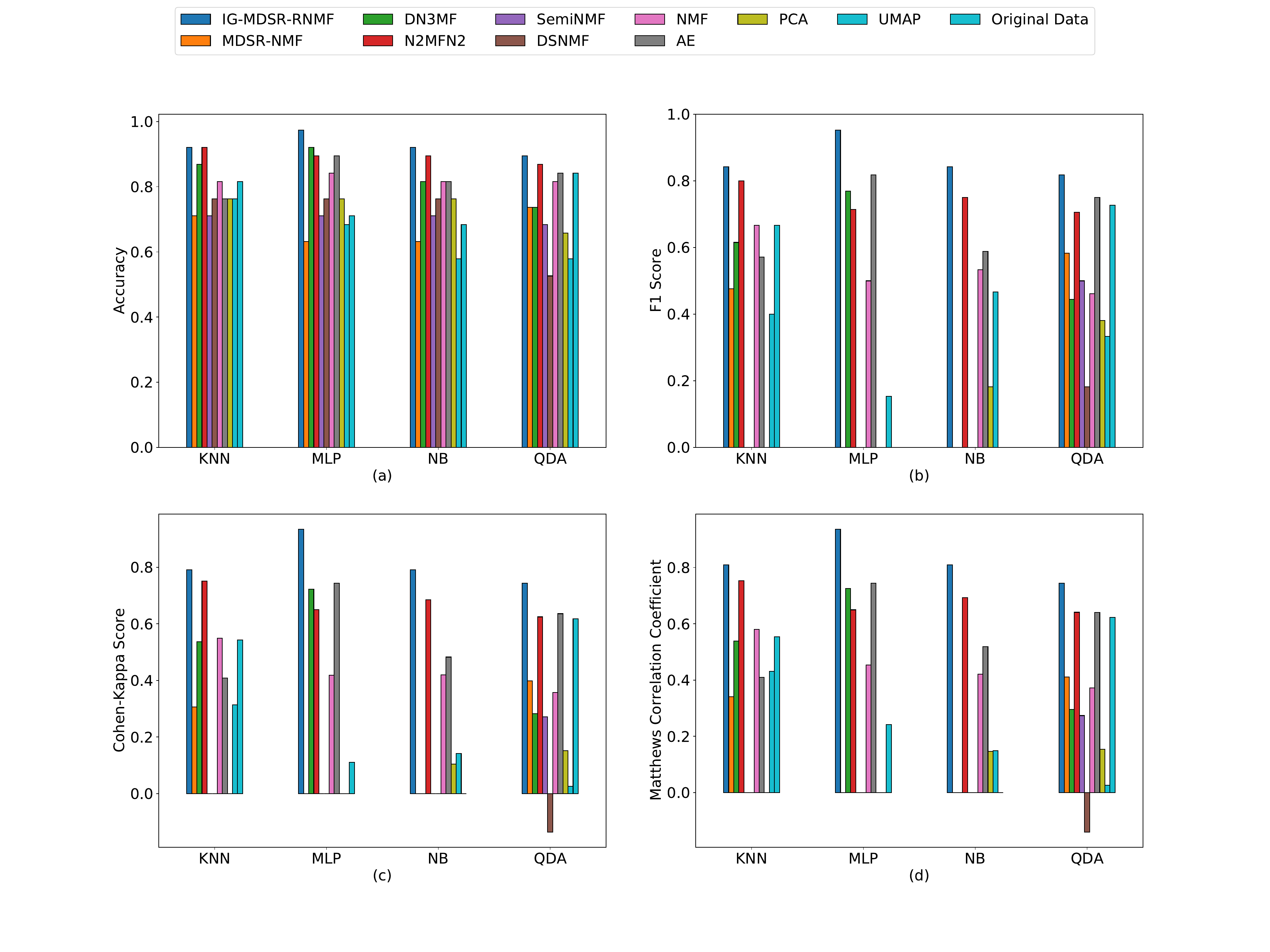}
\caption{Performance scores of the classification algorithms on the dimensionally reduced dataset instances of the GLRC dataset by IG-MDSR-RNMF and nine other dimension reduction techniques along with the original data.}
\label{IG-MDSR-RNMF_classification_GLRC_V33}
\end{figure}
\begin{figure}[ht!]
\centering
\includegraphics[width=\linewidth\iffalse125mm\fi, scale=1.0]{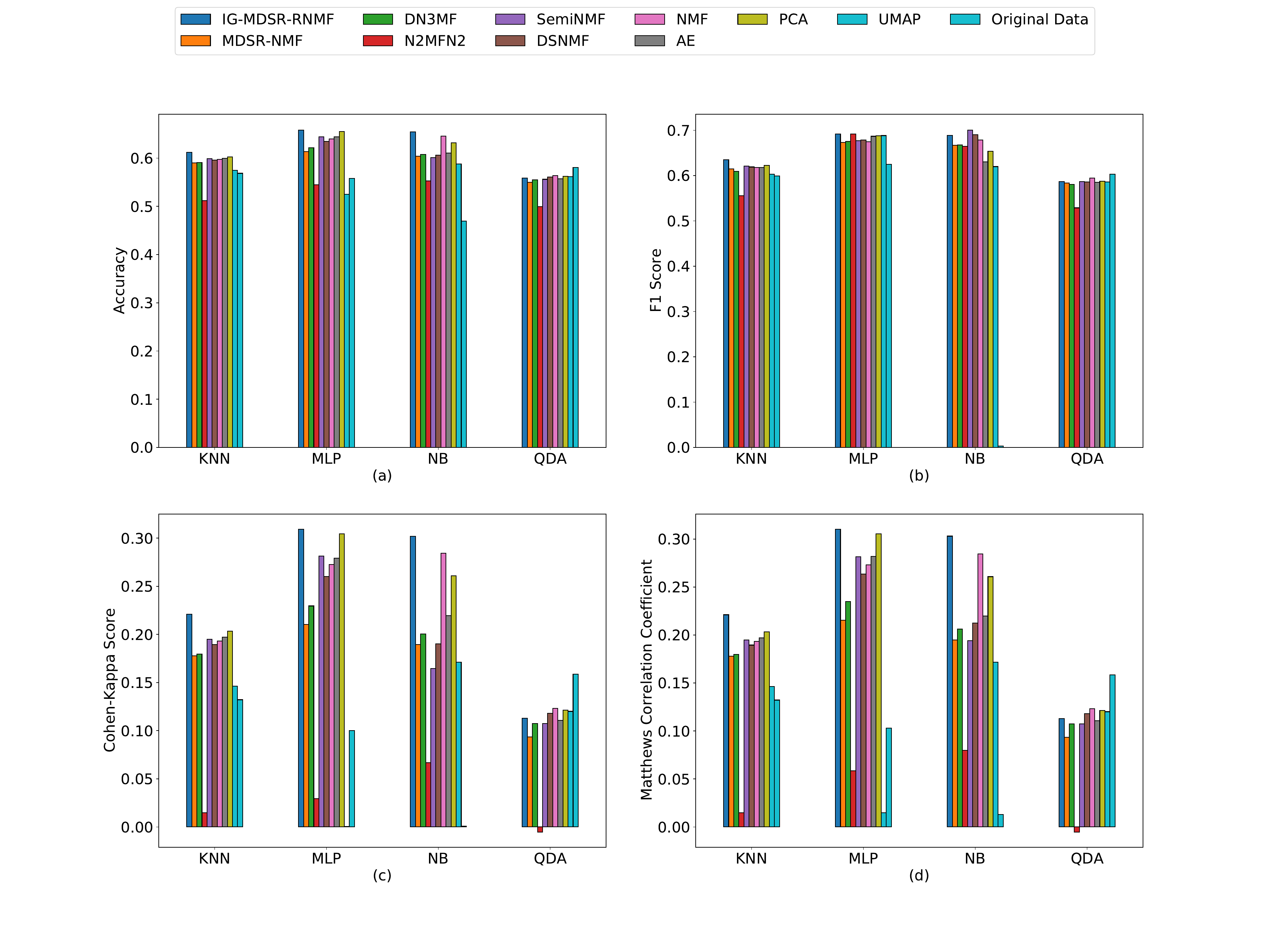}
\caption{Performance scores of the classification algorithms on the dimensionally reduced dataset instances of the ONP dataset by IG-MDSR-RNMF and nine other dimension reduction techniques along with the original data.}
\label{IG-MDSR-RNMF_classification_ONP_V33}
\end{figure}
\begin{figure}[ht!]
\centering
\includegraphics[width=\linewidth\iffalse125mm\fi, scale=1.0]{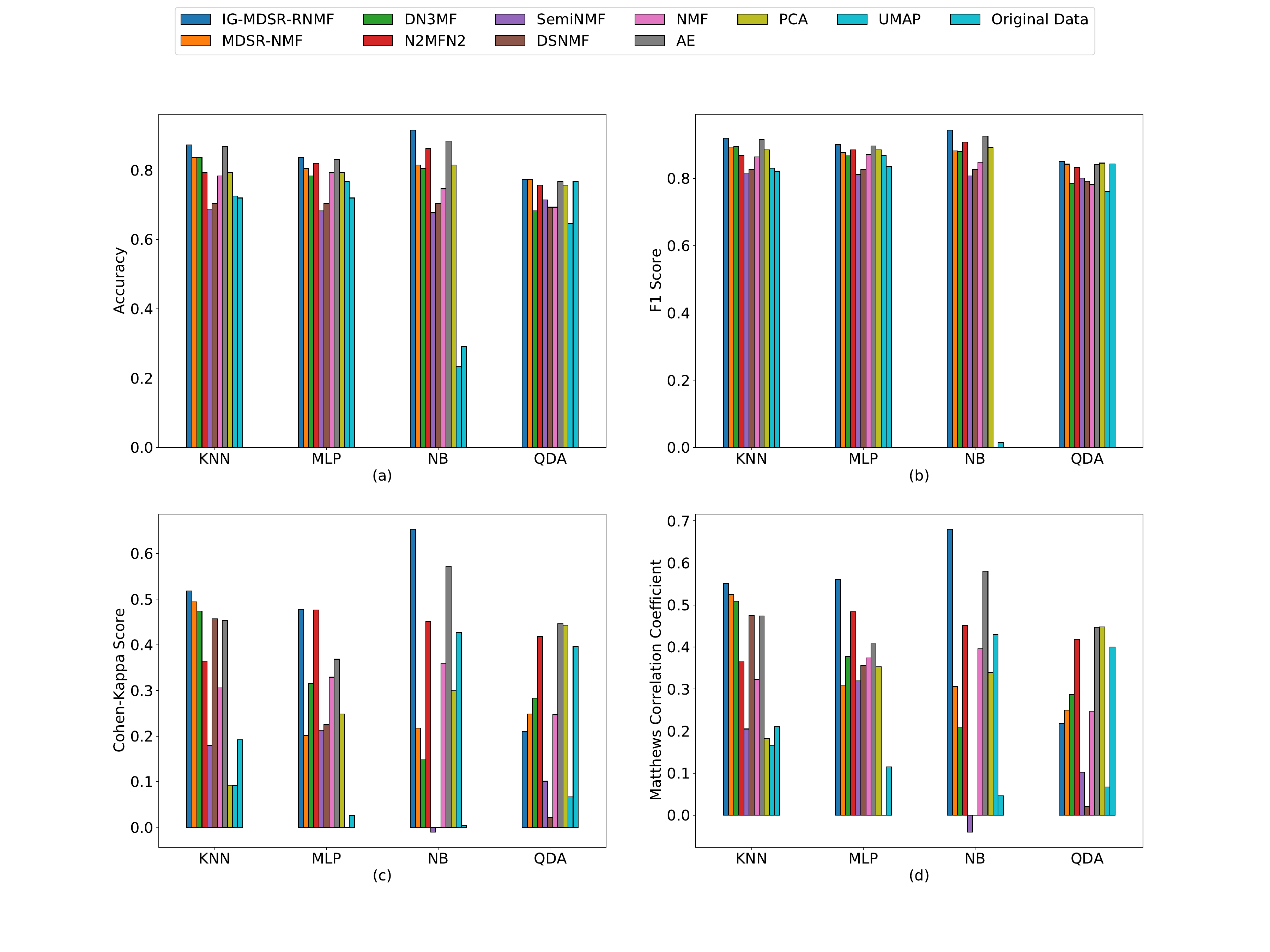}
\caption{Performance scores of the classification algorithms on the dimensionally reduced dataset instances of the PDC dataset by IG-MDSR-RNMF and nine other dimension reduction techniques along with the original data.}
\label{IG-MDSR-RNMF_classification_PDC_V33}
\end{figure}
\begin{figure}[ht!]
\centering
\includegraphics[width=\linewidth\iffalse125mm\fi, scale=1.0]{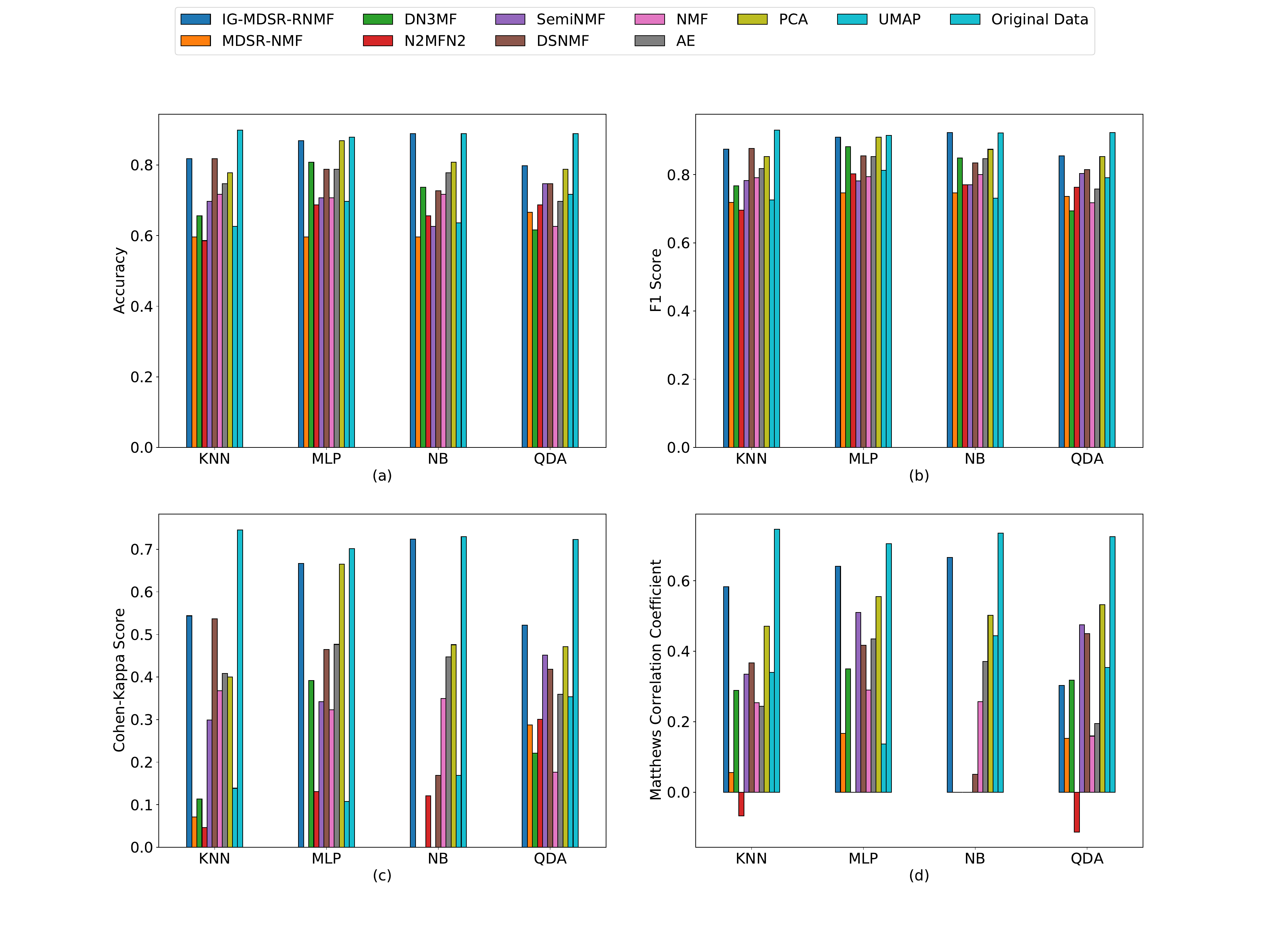}
\caption{Performance scores of the classification algorithms on the dimensionally reduced dataset instances of the SP dataset by IG-MDSR-RNMF and nine other dimension reduction techniques along with the original data.}
\label{IG-MDSR-RNMF_classification_SP_V33}
\end{figure}
\begin{figure}[ht!]
\centering
\includegraphics[width=\linewidth\iffalse125mm\fi, scale=1.0]{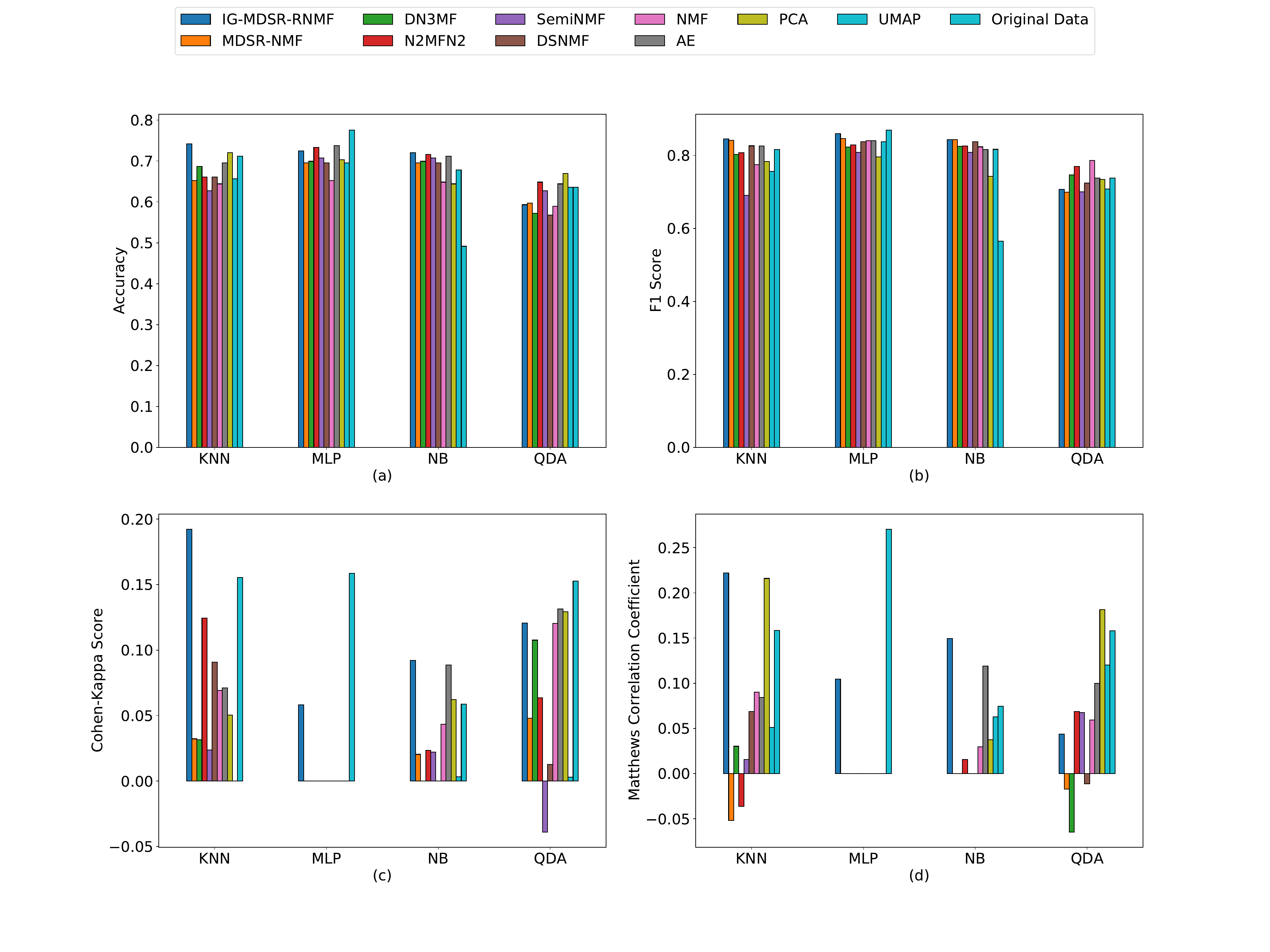}
\caption{Performance scores of the classification algorithms on the dimensionally reduced dataset instances of the MovieLens dataset by IG-MDSR-RNMF and nine other dimension reduction techniques along with the original data.}
\label{IG-MDSR-RNMF_classification_ml-100k_V33}
\end{figure}

Likewise, each reduced dataset $\mathbf{X}_r(\dot{T})$ has been clustered using four well-known clustering techniques: Mini Batch k-Means (MBkM), Balanced Iterative Reducing and Clustering using Hierarchies (BIRCH), Gaussian Mixture Models (GMM) and Fuzzy c-Means (FcM). To assess the quality of the results produced by them, four cluster validity indices have been used: Adjusted Mutual Information score (AMI), Adjusted Rand index (ARI), Jaccard index (JI) and Normalized Mutual Information score (NMI). Hence, for each $\mathbf{X}_r(\dot{T})$, we get a cluster validity score by performing clustering using a clustering algorithm and validating the outcome employing a cluster evaluation index. The same approach has been followed separately for the IG-MDSR-RNMF algorithm too. The performance of IG-MDSR-NMF and IG-MDSR-RNMF has also been compared with the original dataset to justify the efficacy of the dimensionally reduced dataset over the original one and the need for the same.

\begin{figure}[ht!]
\centering
\includegraphics[width=\linewidth\iffalse125mm\fi, scale=1.0]{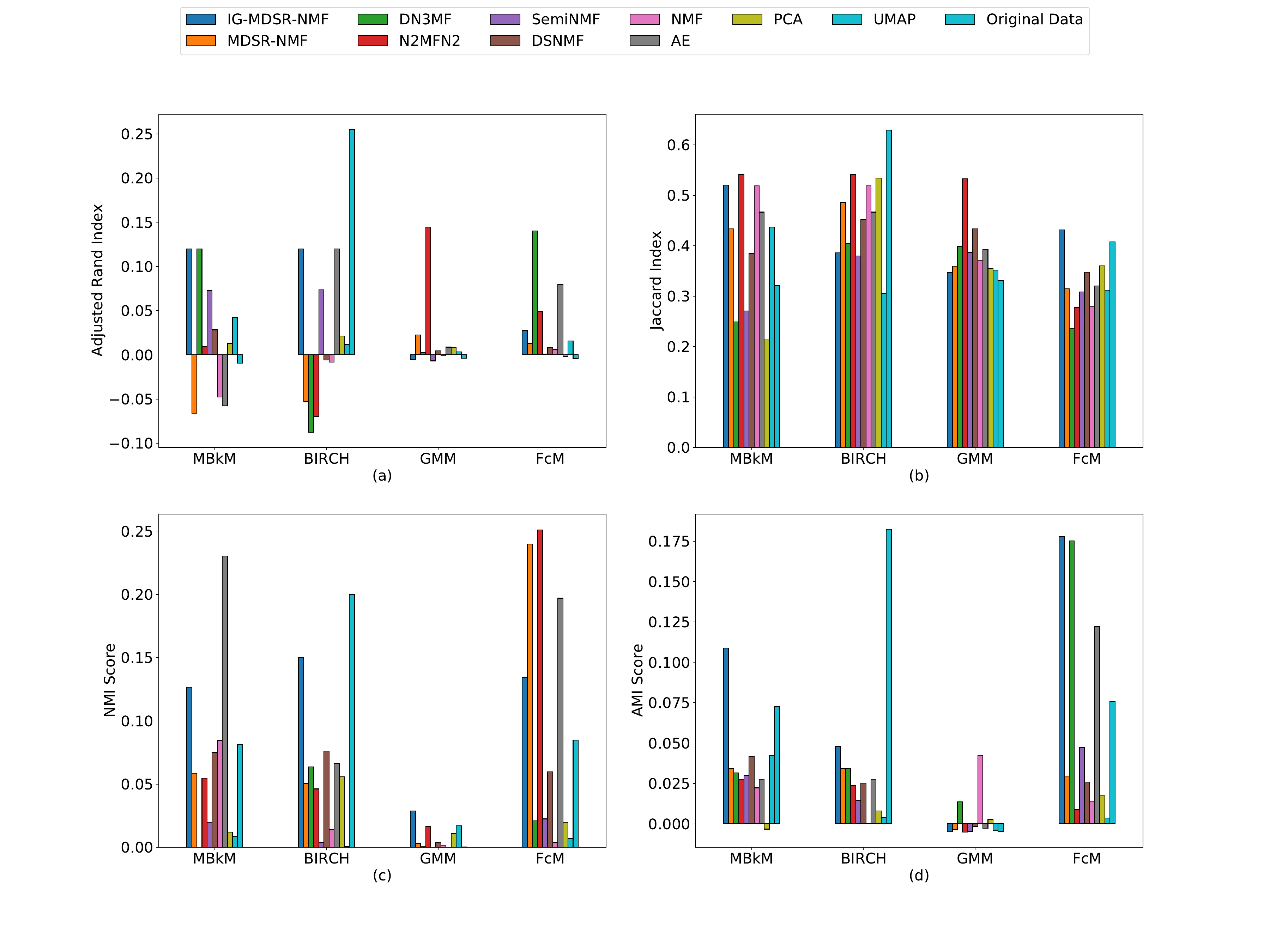}
\caption{Performance scores of the clustering algorithms on the dimensionally reduced dataset instances of the GLRC dataset by IG-MDSR-NMF and nine other dimension reduction techniques along with the original data.}
\label{IG-MDSR-NMF_cluster_GLRC_V31}
\end{figure}
\begin{figure}[ht!]
\centering
\includegraphics[width=\linewidth\iffalse125mm\fi, scale=1.0]{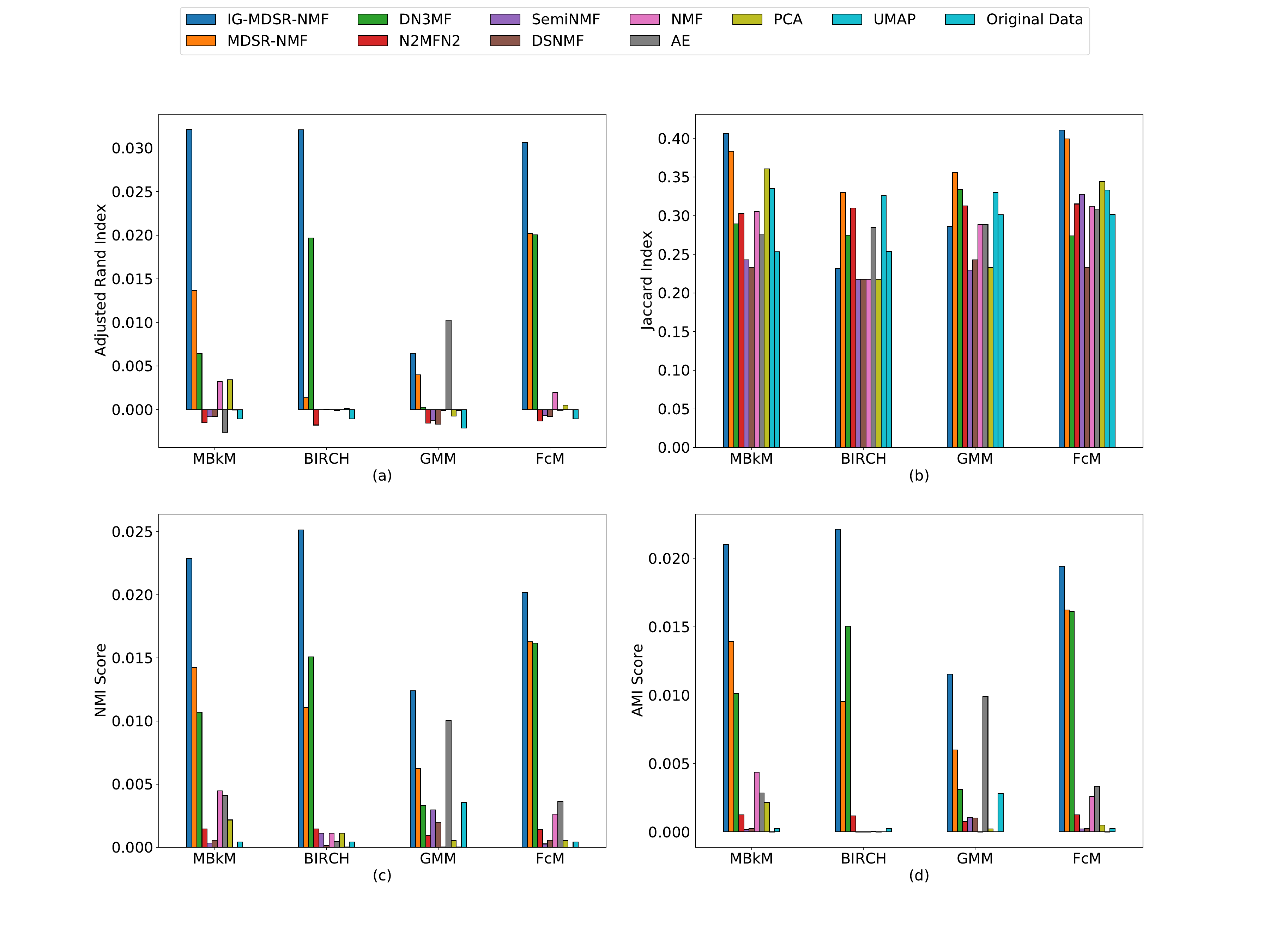}
\caption{Performance scores of the clustering algorithms on the dimensionally reduced dataset instances of the ONP dataset by IG-MDSR-NMF and nine other dimension reduction techniques along with the original data.}
\label{IG-MDSR-NMF_cluster_ONP_V31}
\end{figure}
\begin{figure}[ht!]
\centering
\includegraphics[width=\linewidth\iffalse125mm\fi, scale=1.0]{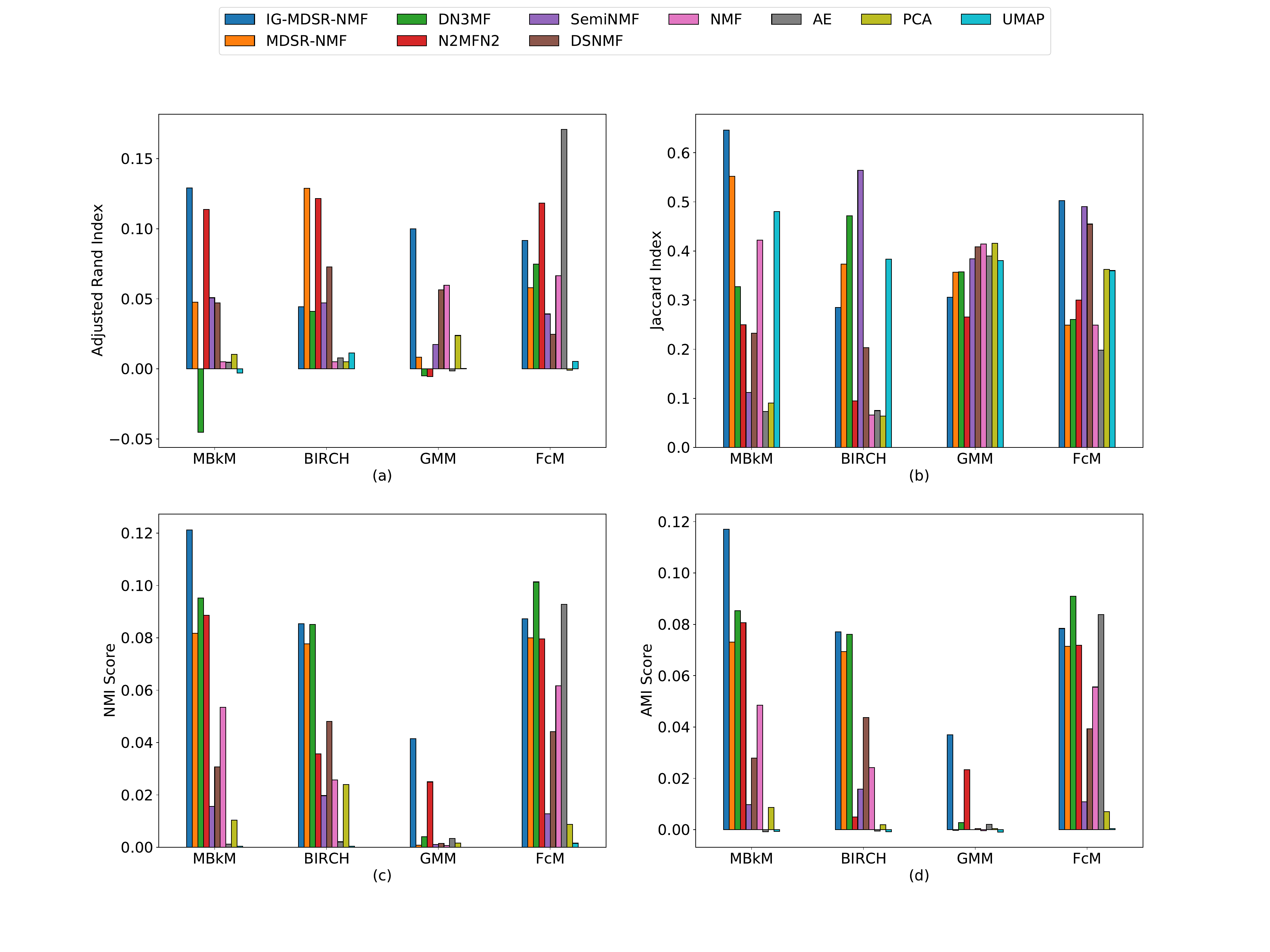}
\caption{Performance scores of the clustering algorithms on the dimensionally reduced dataset instances of the PDC dataset by IG-MDSR-NMF and nine other dimension reduction techniques along with the original data.}
\label{IG-MDSR-NMF_cluster_PDC_V31}
\end{figure}
\begin{figure}[ht!]
\centering
\includegraphics[width=\linewidth\iffalse125mm\fi, scale=1.0]{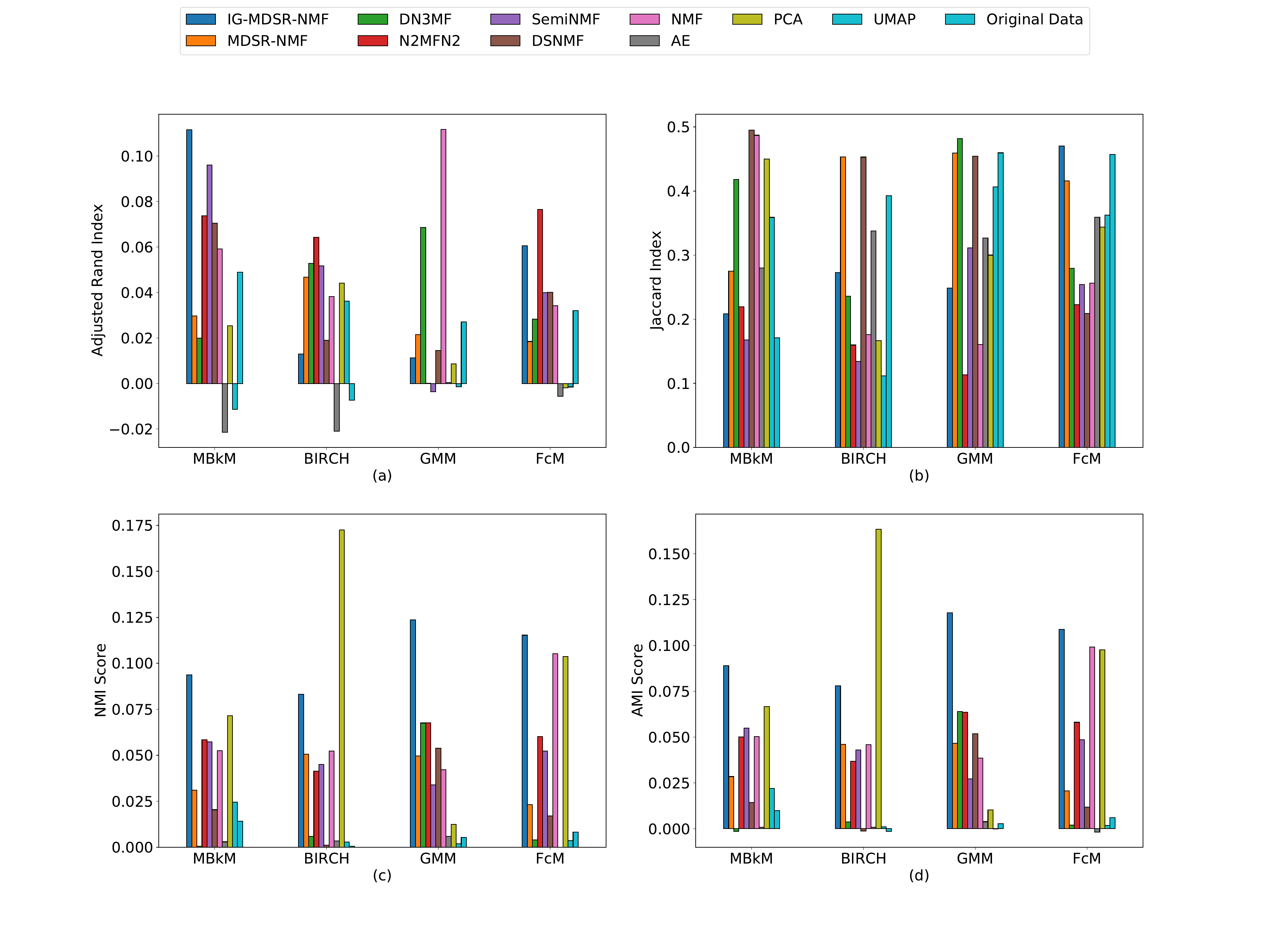}
\caption{Performance scores of the clustering algorithms on the dimensionally reduced dataset instances of the SP dataset by IG-MDSR-NMF and nine other dimension reduction techniques along with the original data.}
\label{IG-MDSR-NMF_cluster_SP_V31}
\end{figure}
\begin{figure}[ht!]
\centering
\includegraphics[width=\linewidth\iffalse125mm\fi, scale=1.0]{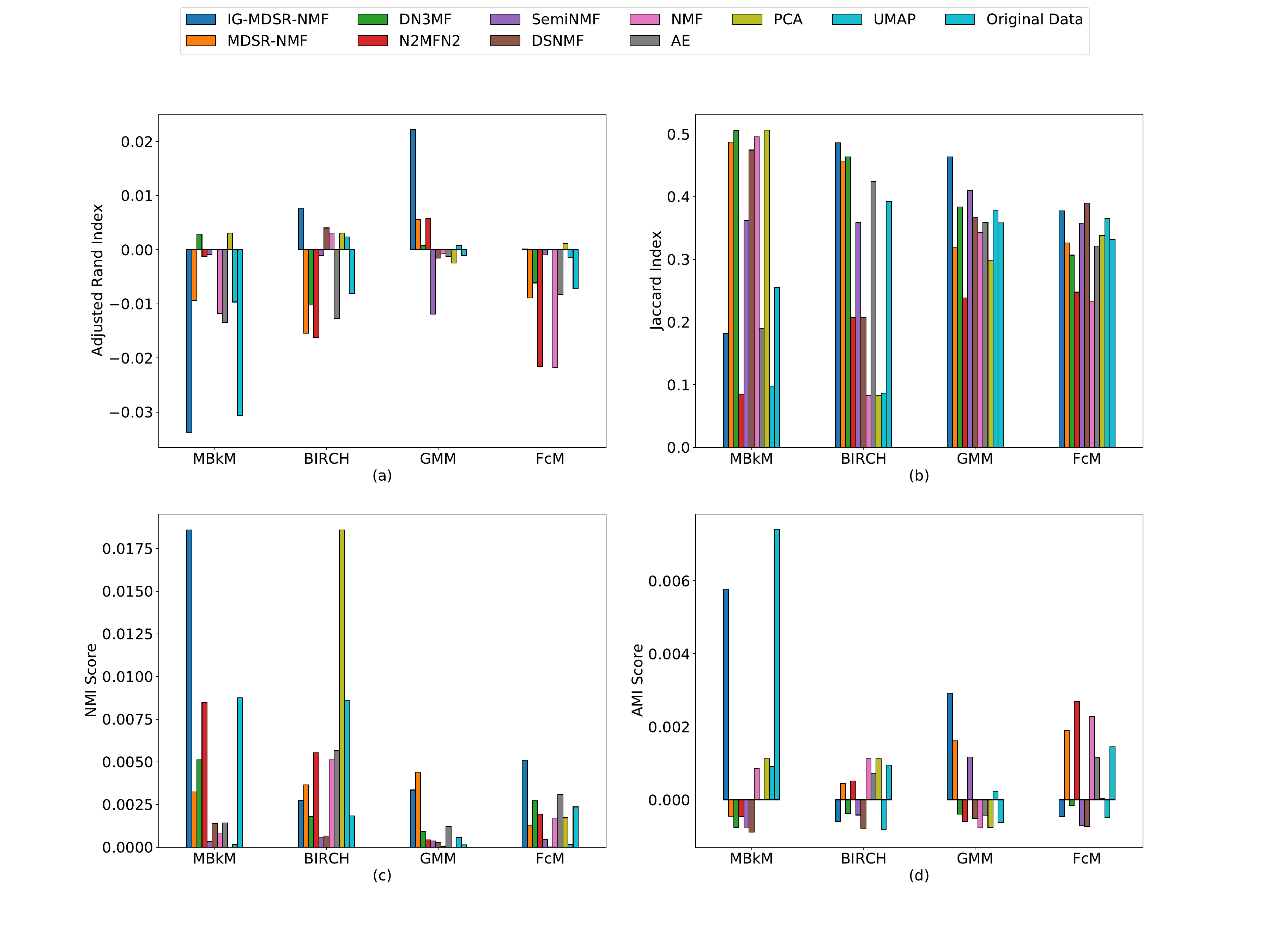}
\caption{Performance scores of the clustering algorithms on the dimensionally reduced dataset instances of the MovieLens dataset by IG-MDSR-NMF and nine other dimension reduction techniques along with the original data.}
\label{IG-MDSR-NMF_cluster_ml-100k_V31}
\end{figure}

\begin{figure}[ht!]
\centering
\includegraphics[width=\linewidth\iffalse125mm\fi, scale=1.0]{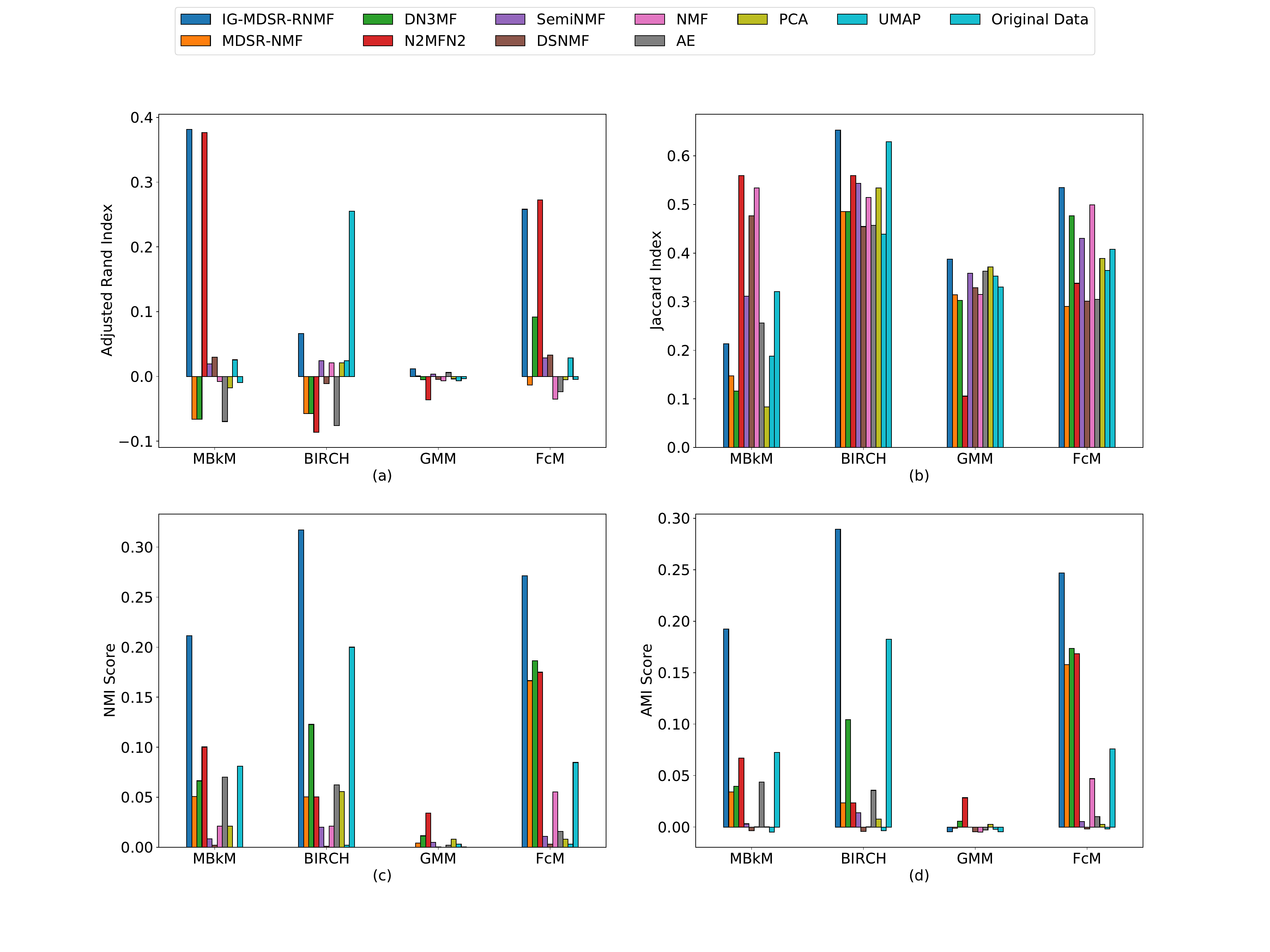}
\caption{Performance scores of the clustering algorithms on the dimensionally reduced dataset instances of the GLRC dataset by IG-MDSR-RNMF and nine other dimension reduction techniques along with the original data.}
\label{IG-MDSR-RNMF_cluster_GLRC_V33}
\end{figure}
\begin{figure}[ht!]
\centering
\includegraphics[width=\linewidth\iffalse125mm\fi, scale=1.0]{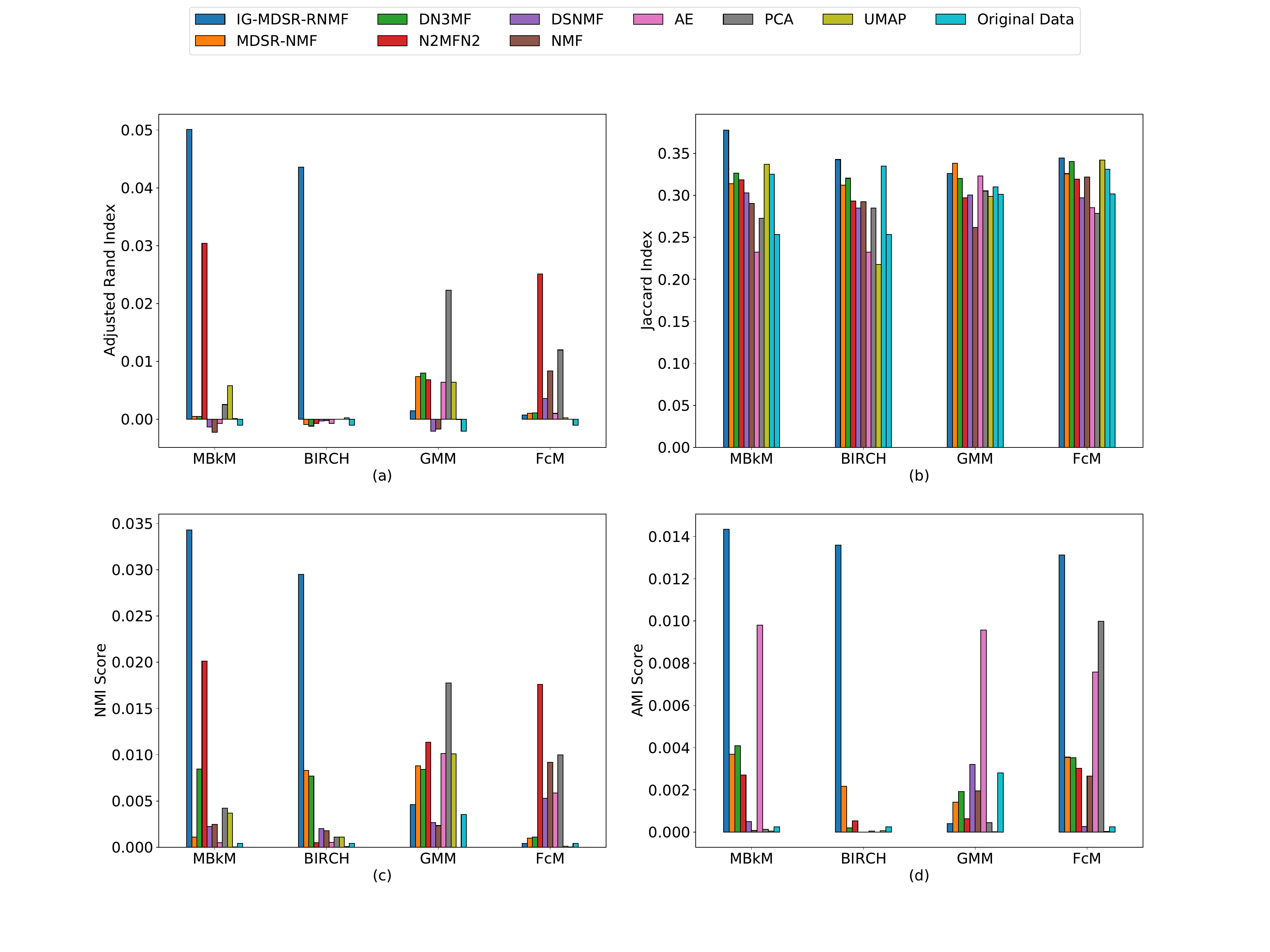}
\caption{Performance scores of the clustering algorithms on the dimensionally reduced dataset instances of the ONP dataset by IG-MDSR-RNMF and nine other dimension reduction techniques along with the original data.}
\label{IG-MDSR-RNMF_cluster_ONP_V33}
\end{figure}
\begin{figure}[ht!]
\centering
\includegraphics[width=\linewidth\iffalse125mm\fi, scale=1.0]{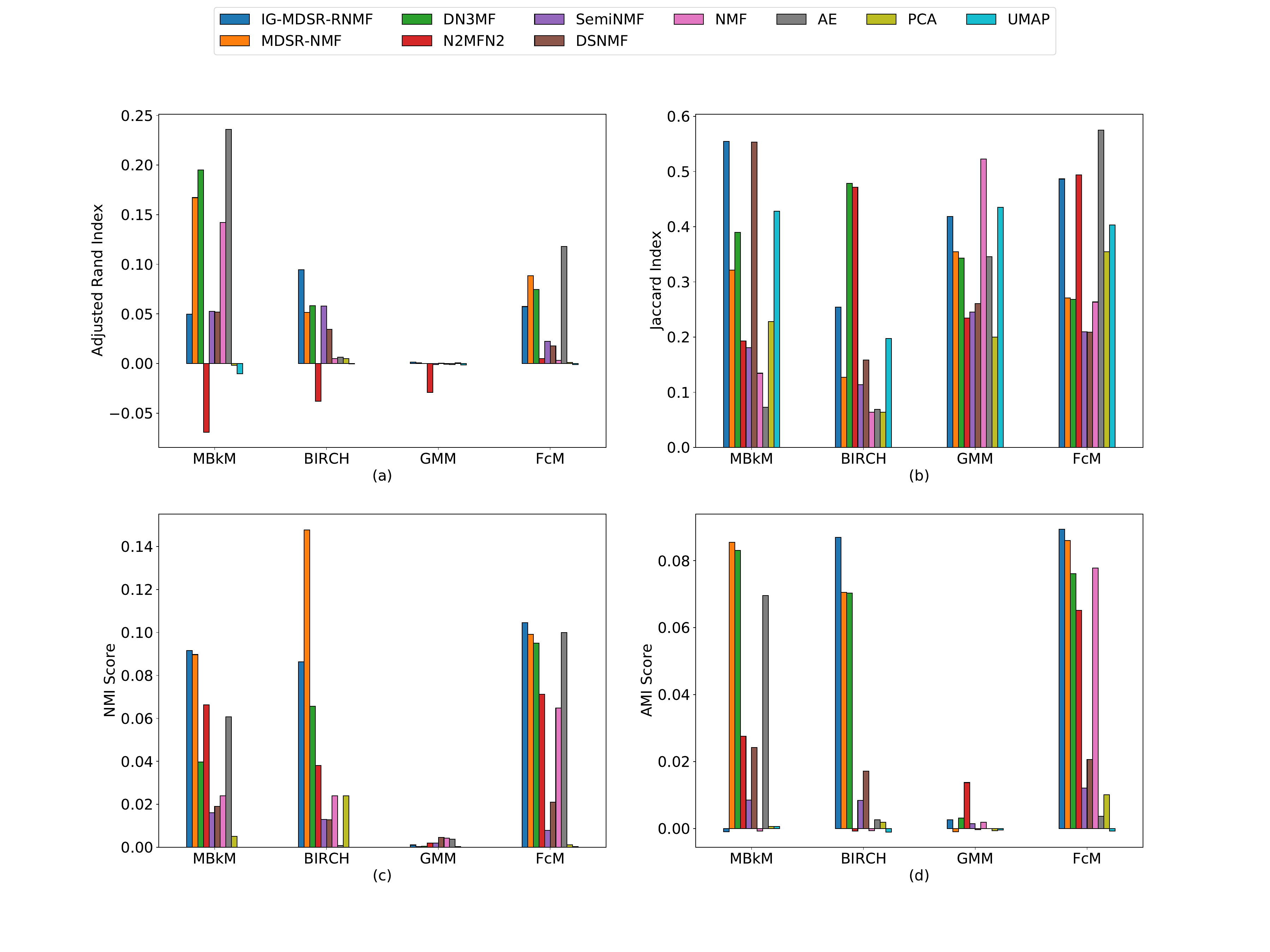}
\caption{Performance scores of the clustering algorithms on the dimensionally reduced dataset instances of the PDC dataset by IG-MDSR-RNMF and nine other dimension reduction techniques along with the original data.}
\label{IG-MDSR-RNMF_cluster_PDC_V33}
\end{figure}
\begin{figure}[ht!]
\centering
\includegraphics[width=\linewidth\iffalse125mm\fi, scale=1.0]{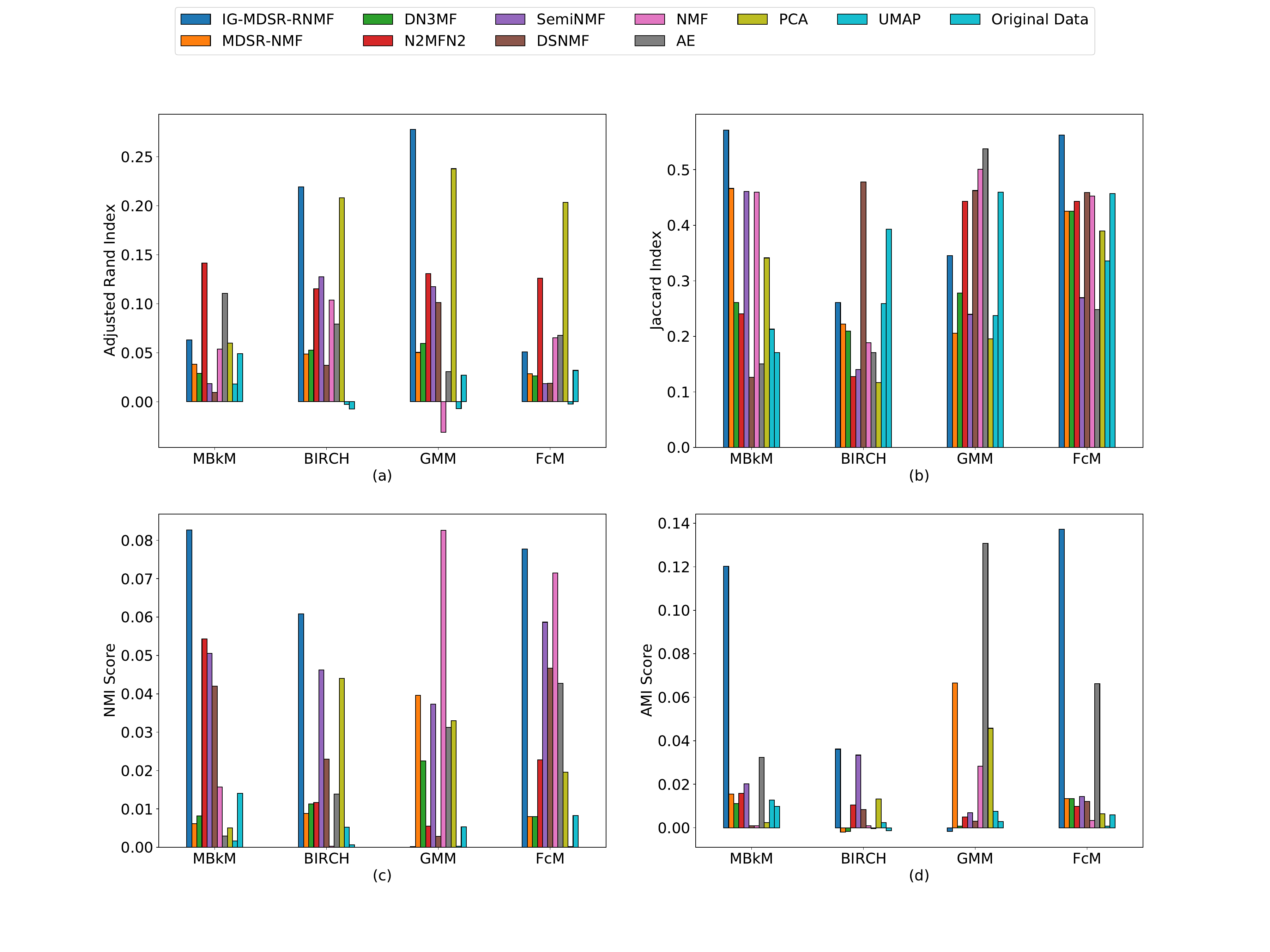}
\caption{Performance scores of the clustering algorithms on the dimensionally reduced dataset instances of the SP dataset by IG-MDSR-RNMF and nine other dimension reduction techniques along with the original data.}
\label{IG-MDSR-RNMF_cluster_SP_V33}
\end{figure}
\begin{figure}[ht!]
\centering
\includegraphics[width=\linewidth\iffalse125mm\fi, scale=1.0]{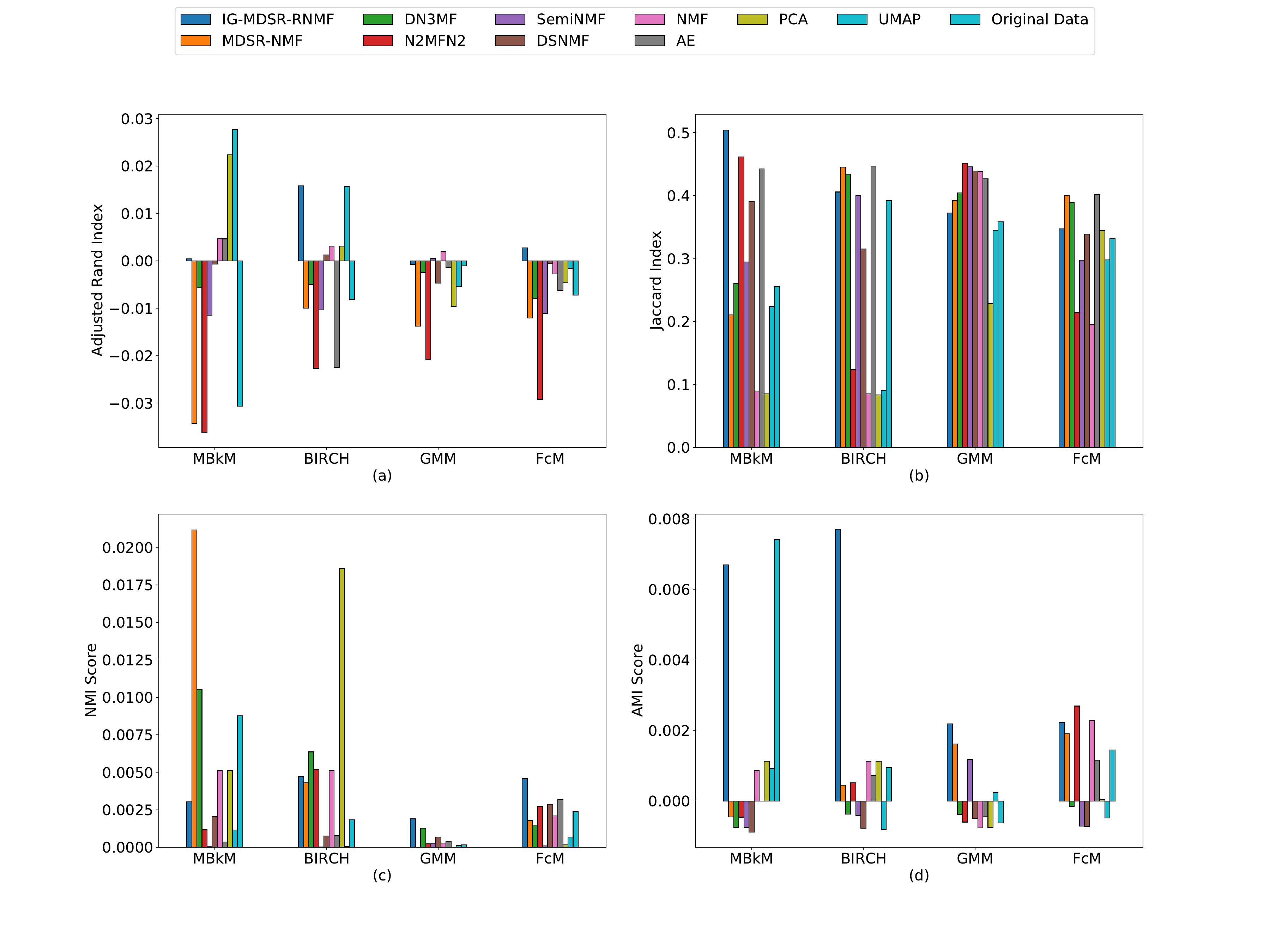}
\caption{Performance scores of the clustering algorithms on the dimensionally reduced dataset instances of the MovieLens dataset by IG-MDSR-RNMF and nine other dimension reduction techniques along with the original data.}
\label{IG-MDSR-RNMF_cluster_ml-100k_V33}
\end{figure}

\subsection{Results and  Analysis}
\label{analysis_IG-MDSR-NMF}
The performance of IG-MDSR-NMF/IG-MDSR-RNMF has been presented and justified in two parts. Firstly, the quality of dimension reduction by the models has been evaluated by comparing their ability to preserve the local structure of data. Secondly, the effectiveness of the dimensionally reduced dataset is explored for downstream analyses, like classification and clustering.

\subsubsection{Quantifying the quality of low dimensional embedding}
The quality of low dimensional embedding by IG-MDSR-NMF/IG-MDSR-RNMF has been investigated in two ways, viz., studying the ability to preserve the local structure of data by using trustworthiness metric and by comparing the effectiveness of dimension reduction by classification/cluster performance metrics compared to the original data.

\noindent\textbf{\\Local structure preservation\\\\}
The ability to preserve the local structure of data after dimension reduction by IG-MDSR-NMF and IG-MDSR-RNMF over that of nine other dimension reduction approaches has been computed and compared using the trustworthiness score. Trustworthiness is a metric to measure the extent of local structure retention in the latent space representation of the data with reference to the original data \cite{van2009Learning, Venna2001Neighborhood, pedregosa2011scikit}. The value of trustworthiness lies between $0$ and $1$, and is defined as
\begin{equation}
T(k) = 1 - \frac{2}{nk (2n - 3k - 1)} \sum^n_{i=1}\sum_{j \in \mathcal{N}_{i}^{k}} \max(0, (r(i, j) - k))
\label{ig_trustworthiness}
\end{equation}
Here, for each sample $i$, $\mathcal{N}_{i}^{k}$ is the set of its $k$ nearest neighbours in the output space and every sample $j$ is its $r(i, j)^{th}$ nearest neighbour in the input space. In other words, any unexpected nearest neighbour in the output space is penalised in proportion to their rank in the input space. The higher the trustworthiness score, the better is the low rank representation, i.e., the better the dimension reduction technique is.

We have computed the trustworthiness score of IG-MDSR-NMF along with nine other dimension reduction techniques considered here. The outcome of the same is depicted in the spider/star plot (Figure~\ref{IG-MDSR-NMF_trustworthyness_V31}). There are five axes corresponding to five datasets. The trustworthiness score of a dimension reduction technique for a particular dataset is a point on that axis. Thus, for a dimension reduction technique, there are five points on five axes corresponding to five datasets. These points can be considered as vertices of a polygon. Thus, in Figure~\ref{IG-MDSR-NMF_trustworthyness_V31}, there are ten polygons for ten dimension reduction techniques. The area covered by a polygon justifies the efficacy of a dimension reduction method over all the datasets together. The higher the area, the better is the performance of the algorithm. From the depiction, we can note that IG-MDSR-NMF has beaten other dimension reduction techniques for GLRC, ONP and PDC datasets. The area bounded by the polygon of IG-MDSR-NMF is shown in a shaded colour in Figure~\ref{IG-MDSR-NMF_trustworthyness_V31}, and it can be observed that it is the highest among all the ten polygons. Thus, the quality of low dimensional embedding produced by IG-MDSR-NMF is superior to that produced by the other dimension reduction methods.

The same procedure has also been followed for IG-MDSR-RNMF to assess the performance of the technique in terms of local shape preservation criteria. The outcome of the same is depicted in Figure~\ref{IG-MDSR-RNMF_trustworthyness_V33}. It is clear from the depiction that IG-MDSR-RNMF has a better trustworthiness score compared to others for four out of five datasets. Hence, the area covered by the polygon corresponding to IG-MDSR-RNMF is higher in contrast to other dimension reduction algorithms. Thus, with similar arguments as above, the superiority of the low dimensional embedding produced by IG-MDSR-RNMF is established.

\noindent\textbf{\\Decision making: Comparison with the original data\\\\}
The efficacy of dimension reduction by IG-MDSR-NMF/IG-MDSR-RNMF has been judged by performing classification and clustering on low dimensional embedding produced by them as well as on the original data, and then quantifying the performances using different classification and cluster validity metrics. This study demonstrates why the dimension reduction is necessary highlighting the fact that the usability of the data increases with the low rank representation of the same.

\noindent\textit{\\Classification\\\\}
Figures~\ref{IG-MDSR-NMF_classification_GLRC_V31}-\ref{IG-MDSR-NMF_classification_ml-100k_V31} presents the performance of IG-MDSR-NMF and original data in terms of classification. For the GLRC dataset (Figure~\ref{IG-MDSR-NMF_classification_GLRC_V31}), IG-MDSR-NMF generated low rank embedding has outperformed the original data for all four classifiers in terms of MCC score; the count with respect to ACC and FS is three out of four and for CKS the count is two. For ONP and MovieLens datasets, for all four performance metrics, IG-MDSR-NMF has performed better than the original dataset for three out of four classification algorithms (Figures~\ref{IG-MDSR-NMF_classification_ONP_V31}, \ref{IG-MDSR-NMF_classification_ml-100k_V31}). In terms of ACC and FS, for the PDC dataset (Figure~\ref{IG-MDSR-NMF_classification_PDC_V31}), the scoreline favouring IG-MDSR-NMF is four out of four. For CKS and MCC metrics, the count is three and two respectively. In the case of the SP dataset, the performance metric of original data is better than the low rank embedding produced by IG-MDSR-NMF on all occasions (Figure~\ref{IG-MDSR-NMF_classification_SP_V31}).

The classification performance comparison of the embedding produced by IG-MDSR-RNMF and the original data have been portrayed in Figures~\ref{IG-MDSR-RNMF_classification_GLRC_V33}-\ref{IG-MDSR-RNMF_classification_ml-100k_V33}. IG-MDSR-RNMF has outnumbered the original dataset in terms of all four classifiers and all four classification performance metrics for the GLRC dataset (Figure~\ref{IG-MDSR-RNMF_classification_GLRC_V33}). This number for the ONP dataset (Figure~\ref{IG-MDSR-RNMF_classification_ONP_V33}) is three out of four for all four metrics and for the MovieLens dataset (Figure~\ref{IG-MDSR-RNMF_classification_ml-100k_V33}) the similar value is two out of four. In terms of ACC and FS metrics, for the PDC dataset, IG-MDSR-RNMF has outperformed the original dataset for all four classifiers, and in the case of CKS and MCC metrics, the number is three (Figure~\ref{IG-MDSR-RNMF_classification_PDC_V33}). For the SP dataset, IG-MDSR-RNMF has outnumbered the original dataset only once for ACC and FS metric (Figure~\ref{IG-MDSR-RNMF_classification_SP_V33}).

Thus, it is evident that in most of the cases, both IG-MDSR-NMF and IG-MDSR-RNMF - projected data have performed better than the original data in terms of classification. This justifies the need for dimension reduction along with the ability to produce low rank embedding preserving elemental characteristics of data.

\noindent\textit{\\Clustering\\\\}
The performance comparison of clustering done on the low dimensional embedding produced by IG-MDSR-NMF and the original data has been illustrated in Figures~\ref{IG-MDSR-NMF_cluster_GLRC_V31}-\ref{IG-MDSR-NMF_cluster_ml-100k_V31}. For the GLRC dataset (Figure~\ref{IG-MDSR-NMF_cluster_GLRC_V31}), for JI and NMI cluster validity indexes, IG-MDSR-NMF has performed better than the original data for three out of four clustering algorithms. For ARI and AMI metrics, the performance score is two out of four in favour of IG-MDSR-NMF. In the case of the ONP dataset (Figure~\ref{IG-MDSR-NMF_cluster_ONP_V31}), IG-MDSR-NMF has outplayed original data for all four clustering techniques in terms of ARI, NMI and AMI metrics. While measuring the performance in terms of JI, IG-MDSR-NMF has scored better for two clustering techniques. For PDC (Figure~\ref{IG-MDSR-NMF_cluster_PDC_V31}) and SP (Figure~\ref{IG-MDSR-NMF_cluster_SP_V31}) datasets, the number of times favouring IG-MDSR-NMF in terms of NMI and AMI metrics is four out of four, whereas for JI, IG-MDSR-NMF has performed better in two out of four clustering algorithms. When measuring the performance in terms of ARI, IG-MDSR-NMF has outperformed the original data for one clustering technique for the PDC dataset and three clustering techniques for the SP dataset. For the MovieLens dataset, the performance score against four clustering methods in favour of IG-MDSR-NMF in terms of ARI, JI, NMI and AMI metrics are two, three, four and one respectively (Figure~\ref{IG-MDSR-NMF_cluster_ml-100k_V31}).

Figures~\ref{IG-MDSR-RNMF_cluster_GLRC_V33}-\ref{IG-MDSR-RNMF_cluster_ml-100k_V33} showcase the performance comparisons of IG-MDSR-RNMF and original data in terms of clustering. For the GLRC dataset, IG-MDSR-RNMF has scored higher thrice among four clustering algorithms in terms of ARI, JI and NMI metrics (Figure~\ref{IG-MDSR-RNMF_cluster_GLRC_V33}). Whereas, in the case of AMI metric IG-MDSR-RNMF has performed better in all four cases. Similarly, for the ONP dataset (Figure~\ref{IG-MDSR-RNMF_cluster_ONP_V33}), IG-MDSR-RNMF has scored better for all four clustering algorithms in terms of ARI, JI and NMI metrics. On the other hand, for the AMI metric IG-MDSR-RNMF has outperformed the original data thrice. For the PDC dataset, the performance statistics favouring IG-MDSR-RNMF against the original data are one, one, four and three in terms of ARI, JI, NMI and AMI metrics (Figure~\ref{IG-MDSR-RNMF_cluster_PDC_V33}) respectively. Similarly, embedding produced by IG-MDSR-RNMF on the MovieLens dataset (Figure~\ref{IG-MDSR-RNMF_cluster_ml-100k_V33}) has outperformed the original embedding for all four clustering methods in terms of ARI and JI metrics, while the same count is three in terms of NMI and AMI metrics. For the SP dataset, the performance of IG-MDSR-RNMF, in terms of ARI, is better for all four clustering techniques, whereas for the NMI and AMI metrics, the performance is better thrice, and for the JI metric the same count is two (Figure~\ref{IG-MDSR-RNMF_cluster_SP_V33}).

Hence, it is established in terms of clustering performance that the low rank embedding generated by both IG-MDSR-NMF and IG-MDSR-RNMF are much better at preserving the fundamental properties of the original data.

\subsubsection{Downstream analyses: Comparison with other models}
By performing classification and clustering on the low dimensional embedding generated by IG-MDSR-NMF/IG-MDSR-RNMF and also that generated by the other nine dimension reduction techniques, the effectiveness of dimension reduction has been assessed. Several metrics measuring classification and cluster performances have been used to quantify the same. This part of the experimentation aims to determine the superiority of dimension reduction by IG-MDSR-NMF/IG-MDSR-RNMF using different types of classification and clustering algorithms.

\noindent\textbf{\\Classification\\\\}
While working with the IG-MDSR-NMF model for classification, the GLRC dataset has been projected to an altered feature space with reduced dimensions of $r=83$ ($f=0.12$), $r=101$ ($f=0.145$), $r=108$ ($f=0.155$) and $r=115$ ($f=0.165$). Similarly, the projected feature space dimension for the ONP dataset is $r=11$ ($f=0.19$), $r=14$ ($f=0.24$), $r=17$ ($f=0.29$), $r=33$ ($f=0.56$), for the PDC dataset, $r=154$ ($f=0.205$), $r=222$ ($f=0.295$), for the SP dataset, $r=13$ ($f=0.41$) and for the MovieLens dataset, $r=319$ ($f=0.19$), $r=344$ ($f=0.205$). The outcome has been depicted by Figures~\ref{IG-MDSR-NMF_classification_GLRC_V31}-\ref{IG-MDSR-NMF_classification_ml-100k_V31}.

For four classification techniques, IG-MDSR-NMF has always achieved the highest accuracy score for GLRC (Figure~\ref{IG-MDSR-NMF_classification_GLRC_V31}(a)), PDC (Figure~\ref{IG-MDSR-NMF_classification_PDC_V31}(a)) and SP (Figure~\ref{IG-MDSR-NMF_classification_SP_V31}(a)) datasets, three times for the MovieLens (Figure~\ref{IG-MDSR-NMF_classification_ml-100k_V31}(a)) dataset and two times for the ONP (Figure~\ref{IG-MDSR-NMF_classification_ONP_V31}(a)) dataset. When measuring classification performance using the F1 score, IG-MDSR-NMF has achieved the highest score in all four classification techniques for the PDC (Figure~\ref{IG-MDSR-NMF_classification_PDC_V31}(b)) dataset, three times for the GLRC (Figure~\ref{IG-MDSR-NMF_classification_GLRC_V31}(b)), MovieLens (Figure~\ref{IG-MDSR-NMF_classification_ml-100k_V31}(b)) and SP (Figure~\ref{IG-MDSR-NMF_classification_SP_V31}(b)) datasets, and twice for the ONP (Figure~\ref{IG-MDSR-NMF_classification_ONP_V31}(b)) dataset. IG-MDSR-NMF has surpassed the others in terms of Cohen-Kappa score on the GLRC (Figure~\ref{IG-MDSR-NMF_classification_GLRC_V31}(c)) and MovieLens (Figure~\ref{IG-MDSR-NMF_classification_ml-100k_V31}(c)) datasets using three of the four classification techniques, two for ONP (Figure~\ref{IG-MDSR-NMF_classification_ONP_V31}(c)), PDC (Figure~\ref{IG-MDSR-NMF_classification_PDC_V31}(c)) and SP (Figure~\ref{IG-MDSR-NMF_classification_SP_V31}(c)) datasets. When the Matthews Correlation Coefficient is used as the classification performance indicator, IG-MDSR-NMF has outperformed the other dimension reduction algorithms three out of four times for the GLRC (Figure~\ref{IG-MDSR-NMF_classification_GLRC_V31}(d)), PDC (Figure~\ref{IG-MDSR-NMF_classification_PDC_V31}(d)) and MovieLens (Figure~\ref{IG-MDSR-NMF_classification_ml-100k_V31}(d)) datasets, and twice for the ONP (Figure~\ref{IG-MDSR-NMF_classification_ONP_V31}(d)) and SP (Figure~\ref{IG-MDSR-NMF_classification_SP_V31}(d)) datasets.

The GLRC dataset has been projected to a modified feature space with a lower dimension of $r=118$ ($f=0.17$) while working with the IG-MDSR-RNMF model for classification. The low rank feature space dimension for the ONP dataset is $r=24$ ($f=0.41$), for the PDC dataset, $r=150$ ($f=0.2$), $r=327$ ($f=0.435$), for the SP dataset, $r=9$ ($f=0.285$), $r=10$ ($f=0.315$), and for the MovieLens dataset, $r=252$ ($f=0.15$), $r=361$ ($f=0.215$), $r=470$ ($f=0.28$). Figures ~\ref{IG-MDSR-RNMF_classification_GLRC_V33}-\ref{IG-MDSR-RNMF_classification_ml-100k_V33} illustrate the outcome.

For four classification techniques, IG-MDSR-RNMF has produced the best accuracy score four times out of four for the GLRC (Figure~\ref{IG-MDSR-RNMF_classification_GLRC_V33}(a)), PDC (Figure~\ref{IG-MDSR-RNMF_classification_PDC_V33}(a)) and SP (Figure~\ref{IG-MDSR-RNMF_classification_SP_V33}(a)) datasets, three times for ONP (Figure~\ref{IG-MDSR-RNMF_classification_ONP_V33}(a)) and twice for MovieLens (Figure~\ref{IG-MDSR-RNMF_classification_ml-100k_V33}(a)). When we use the F1 score to quantify classification performance, IG-MDSR-RNMF has received the highest score in four of the four classification procedures for the GLRC (Figure~\ref{IG-MDSR-RNMF_classification_GLRC_V33}(b)) and PDC (Figure~\ref{IG-MDSR-RNMF_classification_PDC_V33}(b)) datasets, three times for the MovieLens (Figure~\ref{IG-MDSR-RNMF_classification_ml-100k_V33}(b)) and SP (Figure~\ref{IG-MDSR-RNMF_classification_SP_V33}(b)) datasets, and twice for the ONP (Figure~\ref{IG-MDSR-RNMF_classification_ONP_V33}(b)) dataset. For the GLRC (Figure~\ref{IG-MDSR-RNMF_classification_GLRC_V33}(c)) and SP (Figure~\ref{IG-MDSR-RNMF_classification_SP_V33}(c)) datasets, IG-MDSR-RNMF has outscored the others in terms of Cohen-Kappa score across all four classification methods. This value is three for the ONP (Figure~\ref{IG-MDSR-RNMF_classification_ONP_V33}(c)), PDC (Figure~\ref{IG-MDSR-RNMF_classification_PDC_V33}(c)) and MovieLens (Figure~\ref{IG-MDSR-RNMF_classification_ml-100k_V33}(c)) datasets. IG-MDSR-RNMF has beaten the other dimension reduction algorithms four times out of four for the GLRC (Figure~\ref{IG-MDSR-RNMF_classification_GLRC_V33}(d)) dataset and three times for the ONP (Figure~\ref{IG-MDSR-RNMF_classification_ONP_V33}(d)), PDC (Figure~\ref{IG-MDSR-RNMF_classification_PDC_V33}(d)), MovieLens (Figure~\ref{IG-MDSR-RNMF_classification_ml-100k_V33}(d)) and SP (Figure~\ref{IG-MDSR-RNMF_classification_SP_V33}(d)) datasets when the classification performance evaluator is Matthews Correlation Coefficient.

The preceding description clearly shows that in the majority of situations, for all the five datasets and four types of classifiers, the accuracy score of the transformed dataset using IG-MDSR-NMF/IG-MDSR-RNMF has surpassed the others. Accuracy quantifies how often the model is correct. Together with accuracy, we have computed the F1 score, i.e., the harmonic mean of precision and recall. Precision may be thought of as a measure of quality, i.e., the correctness of positive predictions, whereas recall attempts to determine if the model can discover all the instances of the positive class, i.e., recall being a measure of quantity. The F1 score combines the traits of accuracy and recall, making it a stronger metric when the class distribution is uneven. Figures~\ref{IG-MDSR-NMF_classification_GLRC_V31}-\ref{IG-MDSR-NMF_classification_ml-100k_V31} and \ref{IG-MDSR-RNMF_classification_GLRC_V33}-\ref{IG-MDSR-RNMF_classification_ml-100k_V33} show that IG-MDSR-NMF/IG-MDSR-RNMF has outperformed other models in terms of F1 score in most of the situations. Thus, the supremacy of IG-MDSR-NMF/IG-MDSR-RNMF is justified by its Accuracy and F1 score. On the other hand, the Cohen-Kappa score is a statistical indicator of inter-rater agreement. A higher positive value implies good agreement, whereas zero or lower values imply no agreement, i.e., random labelling. The pictorial illustrations show that IG-MDSR-NMF/IG-MDSR-RNMF has resulted in greater positive Cohen-Kappa scores and outperformed the others in the majority of situations. As a consequence, it is possible to conclude that IG-MDSR-NMF/IG-MDSR-RNMF is able to maintain and learn the inherent qualities of the input, resulting in higher scores. Matthews Correlation Coefficient assesses the quality of binary and multiclass classifications by calculating the agreement between predicted and actual classes, taking into account both true and false positives and negatives. A higher MCC score implies better agreement, which means that the model can preserve the class properties of the original dataset in the altered dataset as well. Figures~\ref{IG-MDSR-NMF_classification_GLRC_V31}-\ref{IG-MDSR-NMF_classification_ml-100k_V31} and \ref{IG-MDSR-RNMF_classification_GLRC_V33}-\ref{IG-MDSR-RNMF_classification_ml-100k_V33} show that IG-MDSR-NMF/IG-MDSR-RNMF has outperformed the other models in terms of MCC score. The preceding discussion demonstrates the advantage of IG-MDSR-NMF/IG-MDSR-RNMF over the other dimension reduction algorithms in terms of both statistical and intrinsic property preservation metrics.

\noindent\textbf{\\Clustering\\\\}
For clustering purposes with the IG-MDSR-NMF model, the GLRC dataset has been projected to a low rank space with reduced dimensions of $r=73$ ($f=0.105$), $r=108$ ($f=0.155$), $r=122$ ($f=0.175$) and $r=136$ ($f=0.195$). The projected feature space dimension for the ONP dataset is $r=6$ ($f=0.105$) for the PDC dataset; $r=169$ ($f=0.225$), $r=274$ ($f=0.365$), $r=316$ ($f=0.42$), for the SP dataset; $r=4$ ($f=0.125$), $r=13$ ($f=0.41$); and for the MovieLens dataset, $r=344$ ($f=0.205$), $r=437$ ($f=0.26$). Figures~\ref{IG-MDSR-NMF_cluster_GLRC_V31}-\ref{IG-MDSR-NMF_cluster_ml-100k_V31} present the outcome.

IG-MDSR-NMF has achieved the highest performance score for the Adjusted Rand index for the ONP (Figure~\ref{IG-MDSR-NMF_cluster_ONP_V31}(a)) dataset thrice among the four clustering approaches considered here. This count is two out of four for the GLRC (Figure~\ref{IG-MDSR-NMF_cluster_GLRC_V31}(a)), PDC (Figure~\ref{IG-MDSR-NMF_cluster_PDC_V31}(a)) and MovieLens (Figure~\ref{IG-MDSR-NMF_cluster_ml-100k_V31}(a)) datasets, with one for the SP (Figure~\ref{IG-MDSR-NMF_cluster_SP_V31}(a)) dataset. When using the Jaccard Index as the cluster validity estimator, IG-MDSR-NMF has outperformed the others in two out of four clustering algorithms on the ONP (Figure~\ref{IG-MDSR-NMF_cluster_ONP_V31}(b)), PDC (Figure~\ref{IG-MDSR-NMF_cluster_PDC_V31}(b)) and MovieLens (Figure~\ref{IG-MDSR-NMF_cluster_ml-100k_V31}(b)) datasets. This value ranks one out of four for the GLRC (Figure~\ref{IG-MDSR-NMF_cluster_GLRC_V31}(b)) and SP (Figure~\ref{IG-MDSR-NMF_cluster_SP_V31}(b)) datasets. When the cluster validity index is Normalized Mutual Information score, IG-MDSR-NMF has outperformed others four out of four times for the ONP (Figure~\ref{IG-MDSR-NMF_cluster_ONP_V31}(c)) dataset, thrice for PDC (Figure~\ref{IG-MDSR-NMF_cluster_PDC_V31}(c)) and SP (Figure~\ref{IG-MDSR-NMF_cluster_SP_V31}(c)) datasets, and twice for GLRC (Figure~\ref{IG-MDSR-NMF_cluster_GLRC_V31}(c)) and MovieLens (Figure~\ref{IG-MDSR-NMF_cluster_ml-100k_V31}(c)) datasets. IG-MDSR-NMF projected transformed space has achieved the highest Adjusted Mutual Information score among the other dimension reduction techniques four times on the ONP (Figure~\ref{IG-MDSR-NMF_cluster_ONP_V31}(d)) dataset for all the four clustering algorithms. The count is three for the GLRC (Figure~\ref{IG-MDSR-NMF_cluster_GLRC_V31}(d)), PDC (Figure~\ref{IG-MDSR-NMF_cluster_PDC_V31}(d)) and SP (Figure~\ref{IG-MDSR-NMF_cluster_SP_V31}(d)) datasets. In the MovieLens (Figure~\ref{IG-MDSR-NMF_cluster_ml-100k_V31}(d)) dataset, this value is two out of four.

To analyze the effectiveness of the IG-MDSR-RNMF model for clustering, the GLRC dataset has been projected to an altered feature space with lower dimensions of $r=90$ ($f=0.13$), $r=129$ ($f=0.185$) and $r=143$ ($f=0.205$). The low rank feature space dimensions for the ONP dataset are $r=5$ ($f=0.1$), $r=23$ ($f=0.39$), $r=35$ ($f=0.595$); $r=139$ ($f=0.185$), $r=353$ ($f=0.47$), $r=402$ ($f=0.535$), $r=410$ ($f=0.545$) for the PDC dataset; $r=3$ ($f=0.1$), $r=7$ ($f=0.22$), $r=18$ ($f=0.565$) for the SP dataset; and $r=218$ ($f=0.13$), $r=277$ ($f=0.165$), $r=403$ ($f=0.24$), $r=437$ ($f=0.26$) for the MovieLens dataset. The results are shown in Figures~\ref{IG-MDSR-RNMF_cluster_GLRC_V33}-\ref{IG-MDSR-RNMF_cluster_ml-100k_V33}.

Three of the four clustering algorithms studied here get the best performance score for the Adjusted Rand Index for the GLRC (Figure~\ref{IG-MDSR-RNMF_cluster_GLRC_V33}(a)) dataset using IG-MDSR-RNMF. ONP (Figure~\ref{IG-MDSR-RNMF_cluster_ONP_V33}(a)), PDC (Figure~\ref{IG-MDSR-RNMF_cluster_PDC_V33}(a)), MovieLens (Figure~\ref{IG-MDSR-RNMF_cluster_ml-100k_V33}(a)) and SP (Figure~\ref{IG-MDSR-RNMF_cluster_SP_V33}(a)) datasets have a two-out-of-four count. On the GLRC (Figure~\ref{IG-MDSR-RNMF_cluster_GLRC_V33}(b)) and ONP (Figure~\ref{IG-MDSR-RNMF_cluster_ONP_V33}(b)) datasets, IG-MDSR-RNMF has outperformed the other three clustering algorithms when the Jaccard Index has been used to determine cluster validity. This value is two out of four for the SP (Figure~\ref{IG-MDSR-RNMF_cluster_SP_V33}(b)) dataset and one each for the PDC (Figure~\ref{IG-MDSR-RNMF_cluster_PDC_V33}(b)) and MovieLens (Figure~\ref{IG-MDSR-RNMF_cluster_ml-100k_V33}(b)) datasets. With the Normalised Mutual Information score as the cluster validity metric, IG-MDSR-RNMF has outperformed the other methods thrice for the GLRC (Figure~\ref{IG-MDSR-RNMF_cluster_GLRC_V33}(c)) and SP (Figure~\ref{IG-MDSR-RNMF_cluster_SP_V33}(c)) datasets, and twice for the ONP (Figure~\ref{IG-MDSR-RNMF_cluster_ONP_V33}(c)), PDC (Figure~\ref{IG-MDSR-RNMF_cluster_PDC_V33}(c)), and MovieLens (Figure~\ref{IG-MDSR-RNMF_cluster_ml-100k_V33}(c)) datasets. Among various dimension reduction strategies, IG-MDSR-RNMF projected transformed space has produced the highest Adjusted Mutual Information score three times on the GLRC (Figure~\ref{IG-MDSR-RNMF_cluster_GLRC_V33}(d)), ONP (Figure~\ref{IG-MDSR-RNMF_cluster_ONP_V33}(d)), MovieLens (Figure~\ref{IG-MDSR-RNMF_cluster_ml-100k_V33}(d)) and SP (Figure~\ref{IG-MDSR-RNMF_cluster_SP_V33}(d)) datasets for all the four clustering algorithms that are being studied here. The same count is two for the PDC (Figure~\ref{IG-MDSR-RNMF_cluster_PDC_V33}(d)) dataset.

Adjusted Rand Index (ARI) evaluates the similarity of two data clusterings by considering all pairs of samples and counting whether they are assigned to the same or different clusters in the predicted and true clusterings. Figures~\ref{IG-MDSR-NMF_cluster_GLRC_V31}-\ref{IG-MDSR-NMF_cluster_ml-100k_V31} and \ref{IG-MDSR-RNMF_cluster_GLRC_V33}-\ref{IG-MDSR-RNMF_cluster_ml-100k_V33} show that IG-MDSR-NMF/IG-MDSR-RNMF has outperformed other dimension reduction techniques across five datasets and four clustering algorithms in terms of ARI score. Jaccard Index is used to determine the similarity of two sets. IG-MDSR-NMF/IG-MDSR-RNMF has outperformed the rest in terms of the Jaccard Index. Thus, it may be argued that IG-MDSR-NMF/IG-MDSR-RNMF has learnt the input's fundamental features and mapped them to a low rank representation well. NMI is defined as the normalisation of the Mutual Information score to scale the outcomes in [0,1]. This metric is unadjusted for chance. The AMI score, on the other hand, is invariant to the permutation of the class or cluster label, which means that it is unaffected by the labels' absolute values. Figures~\ref{IG-MDSR-NMF_cluster_GLRC_V31}-\ref{IG-MDSR-NMF_cluster_ml-100k_V31} and \ref{IG-MDSR-RNMF_cluster_GLRC_V33}-\ref{IG-MDSR-RNMF_cluster_ml-100k_V33} show that IG-MDSR-NMF/IG-MDSR-RNMF has outperformed other dimension reduction strategies in terms of both NMI and AMI scores. The improved performance of IG-MDSR-NMF/IG-MDSR-RNMF demonstrates that the low rank representation of the datasets using IG-MDSR-NMF/IG-MDSR-RNMF has been able to maintain the inherent qualities of the original data better than the other approaches considered here.

\subsubsection{Discussion}
IG-MDSR-NMF/IG-MDSR-RNMF model has been evaluated using five real-world datasets. The datasets can be classified into two groups based on their dimensions. ONP, PDC and SP datasets belong to a group that has more samples than features ($m>n'$), whereas GLRC and MovieLens datasets have fewer samples than features ($m<n'$).

Different types of classification techniques have been considered to validate the performance of low rank approximation by the IG-MDSR-NMF/IG-MDSR-RNMF model. For example, KNN is a non-parametric approach; whereas NB algorithm is a probabilistic classifier; MLP is a feed-forward artificial neural network; and QDA is a generative model-based classifier. We have also used two different types of clustering algorithms. MBkM, FcM and GMM are centroid-based clustering algorithms. On the other hand, BIRCH is a hierarchical clustering methodology.

For five datasets, four classification algorithms and four classification performance measures, a total of $5\times4\times4=80$ performance scores are there for each dimension reduction algorithm. It can be seen that IG-MDSR-NMF projected datasets have performed better than the original data on $49$ out of $80$ occasions and the same count for IG-MDSR-RNMF is $50$. On the other hand, IG-MDSR-NMF has secured the highest rating of $57$ times out of $80$ when comparing the performance with other dimension reduction algorithms, which is $66$ for IG-MDSR-RNMF. Thus, the supremacy of IG-MDSR-NMF/IG-MDSR-RNMF over the others is indubitable.

Similar to classification, for clustering also four clustering algorithms and four cluster validity metrics have been used over five datasets to justify the competence of IG-MDSR-NMF/IG-MDSR-RNMF. For clustering, when comparing the performance against the original data, out of $80$ possible cases, IG-MDSR-NMF and IG-MDSR-RNMF have registered better performance 58 and 63 times respectively. The count of supremacy for both IG-MDSR-NMF and IG-MDSR-RNMF with respect to other dimension reduction algorithms stands out to be $47$ out of $80$. In light of the preceding discussion, it is clear that in the majority of circumstances, IG-MDSR-NMF/IG-MDSR-RNMF has proven to be superior to the other dimension reduction approaches considered here.

As previously stated, the datasets on which we have experimented are divided into two sets depending on the connection between the number of samples and attributes. IG-MDSR-NMF/IG-MDSR-RNMF has demonstrated superiority for both types of datasets, proving its ability in dimension reduction, invariance to the relationship between the number of samples and attributes. Furthermore, the value of the reduced dimension is not limited by the number of samples or attributes. These properties distinguish IG-MDSR-NMF/IG-MDSR-RNMF from several other widely used dimension reduction approaches. As a result, it is established that IG-MDSR-NMF/IG-MDSR-RNMF is widely applicable and not limited by input dimension. As a consequence, IG-MDSR-NMF/IG-MDSR-RNMF has outperformed the other nine cutting-edge dimension reduction methods for various classification and clustering approaches on two separate categories of datasets. As the non-negativity restrictions for one of the factors of the original data matrix have been loosened for IG-MDSR-RNMF, the factorization quality has improved, which is reflected in the performance outcome. Thus, the notion of making the coefficient matrix unbound stands out, opening the way for improved latent space representation learning.

The above two modes of experiments establish the superior performance of IG-MDSR-NMF/IG-MDSR-RNMF not only competing with nine other state-of-the-art dimension reduction techniques but also showcasing their ability to preserve the local structure of data. The need for dimension reduction in contrast to working with the original data has also been demonstrated. The novel input-guided design of the architecture distinguishes itself from others through a series of experiments. Moreover, the concept of relaxing the coefficient matrix presents itself as a robust idea of matrix factorization without compromising the latent space representation while enhancing the learning of the same.

\subsection{Convergence Analysis}
\label{convergence_IG-MDSR-NMF}
\begin{figure}[ht!]
    \centering
    \includegraphics[width=1.0\textwidth]{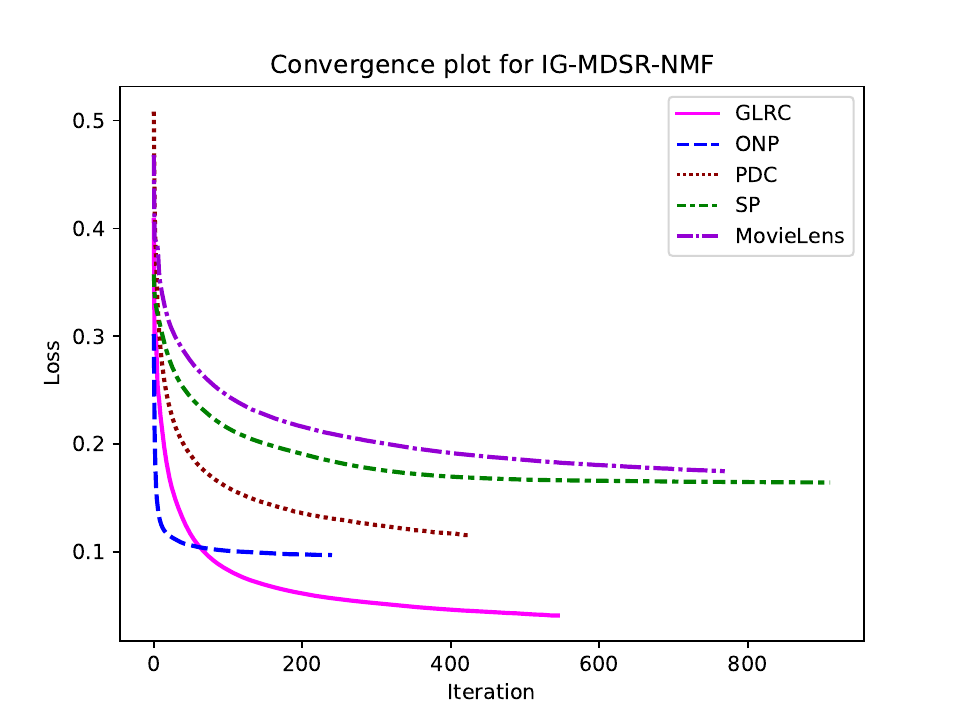}
    \caption{Loss vs. iteration plots of IG-MDSR-NMF for GLRC, ONP, PDC, SP and MovieLens dataset.}
    \label{IG-MDSR-NMF_convergence_V31}
\end{figure}

\begin{figure}[ht!]
    \centering
    \includegraphics[width=1.0\textwidth]{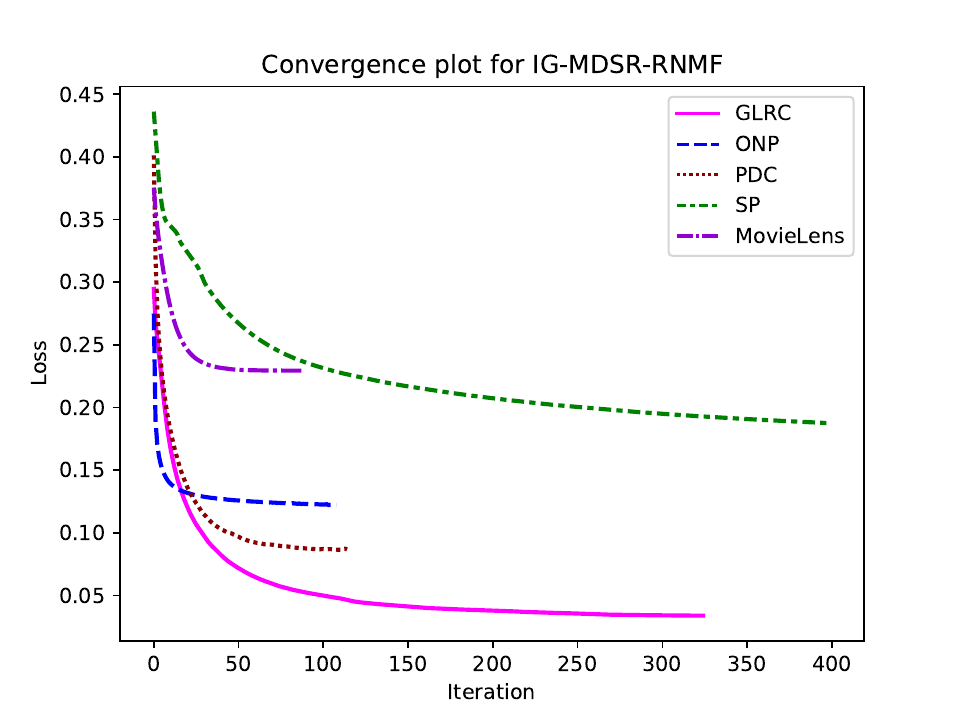}
    \caption{Loss vs. iteration plots of IG-MDSR-RNMF for GLRC, ONP, PDC, SP and MovieLens dataset.}
    \label{IG-MDSR-RNMF_convergence_V33}
\end{figure}

Based on the experimental results, we aim to establish the convergence of the proposed IG-MDSR-NMF/IG-MDSR-RNMF model. The convergence plots for both models are shown in Figures~\ref{IG-MDSR-NMF_convergence_V31} and \ref{IG-MDSR-RNMF_convergence_V33}. The plots illustrate the variation of the cost function $\Phi$ against iteration for all five datasets. Overall, the decreasing nature of the cost over time validates that both models converge. It can also be observed from the plots that the initial cost value for all the datasets starts from a high position and after a few initial epochs, the value of the cost function has almost reached a straight line parallel to the horizontal axis. That is, there are very nominal changes in the cost value. Thus, we can conclude that the models have converged.

\section{Conclusion}
\label{conclusion_IG-MDSR-NMF}
There are numerous techniques to dimensionally reduce a huge dataset with a large number of attributes. In this paper, we have combined the benefits of NMF, a conventional matrix factorization technique, and deep learning for dimension reduction to develop two novel models (IG-MDSR-NMF and IG-MDSR-RNMF) of neural networks. The way how a human being learns a new concept by frequently referring to the original text to maintain the proper direction of learning and enhancing the effectiveness of knowledge gain has inspired the design of IG-MDSR-NMF/IG-MDSR-RNMF. At each step of hierarchical learning, IG-MDSR-NMF/IG-MDSR-RNMF is assisted by the input and hence the models are called ``Input Guided". IG-MDSR-NMF has been constructed in such a manner that it mimics the factorization behaviour of the classic NMF technique. Whereas, IG-MDSR-RNMF introduces an improved type of learning through a modified type of matrix factorization approach called Relaxed-NMF, where only the basis matrix follows the non-negativity criteria. Extensive analysis of the quality of dimension reduction of five popular datasets using IG-MDSR-NMF/IG-MDSR-RNMF has been performed to compare with that of nine other dimension reduction algorithms. These nine dimension reduction techniques include six NMF-based approaches and three conventional dimension reduction algorithms. Preserving the local shape of the original data in the altered space is considered a benchmark of the dimension reduction algorithms. IG-MDSR-NMF/IG-MDSR-RNMF has outclassed other dimension reduction methodologies in terms of local shape preservation. The quality of low rank embedding by IG-MDSR-NMF/IG-MDSR-RNMF has been tested and validated by comparing their performances using both classification and clustering over that of the original dataset, which in turn justifies the need for dimension reduction. During experimentation, a total of four classification algorithms, four classification performance measures, four clustering methods and four cluster validity indexes have been used. The findings also support IG-MDSR-NMF/IG-MDSR-RNMF's superiority over other dimension reduction methodologies evaluated here. Experiments have also been carried out to demonstrate the convergence of IG-MDSR-NMF/IG-MDSR-RNMF.

\textbf{\\\\Acknowledgements.\\}{PD acknowledges Institute of Data Engineering, Analytics and Science Foundation, Technology Innovation Hub, Indian Statistical Institute, Kolkata for providing support and infrastructure to carry out this research work.}
\textbf{\\\\Code availability.\\} The IG-MDSR-NMF/IG-MDSR-RNMF implementations are available at \\https://github.com/duttaprasun/IG-MDSR-NMF.
\textbf{\\\\Author contribution.\\} PD conceived the idea, developed and implemented the methodology, analyzed the results, and made the first draft of the manuscript. RKD provided fruitful suggestions, edited the manuscript, and supervised the work.

\bibliographystyle{unsrt}
\bibliography{references}

\begin{thebibliography}{10}

\bibitem{lee1999learning}
Daniel~D Lee and H~Sebastian Seung.
\newblock {Learning the parts of objects by non-negative matrix factorization}.
\newblock {\em Nature}, 401(6755):788--791, 1999.

\bibitem{lee2001algorithms}
Daniel~D Lee and H~Sebastian Seung.
\newblock {Algorithms for Non-negative Matrix Factorization}.
\newblock In {\em Proceedings of the Advances in Neural Information Processing
  Systems}, pages 556--562, 2001.

\bibitem{trigeorgis2014deep}
George Trigeorgis, Konstantinos Bousmalis, Stefanos Zafeiriou, and Bjoern
  Schuller.
\newblock {A deep semi-nmf model for learning hidden representations}.
\newblock In {\em International conference on machine learning}, pages
  1692--1700. PMLR, 2014.

\bibitem{trigeorgis2016deep}
George Trigeorgis, Konstantinos Bousmalis, Stefanos Zafeiriou, and Bj{\"o}rn~W
  Schuller.
\newblock {A deep matrix factorization method for learning attribute
  representations}.
\newblock {\em IEEE Transactions on Pattern Analysis and Machine Intelligence},
  39(3):417--429, 2016.

\bibitem{ye2018deep}
Fanghua Ye, Chuan Chen, and Zibin Zheng.
\newblock {Deep autoencoder-like nonnegative matrix factorization for community
  detection}.
\newblock In {\em Proceedings of the 27th ACM international conference on
  information and knowledge management}, pages 1393--1402, 2018.

\bibitem{song2015hierarchical}
Hyun~Ah Song, Bo-Kyeong Kim, Thanh~Luong Xuan, and Soo-Young Lee.
\newblock {Hierarchical feature extraction by multi-layer non-negative matrix
  factorization network for classification task}.
\newblock {\em Neurocomputing}, 165:63--74, 2015.

\bibitem{guo2019sparse}
Zhenxing Guo and Shihua Zhang.
\newblock {Sparse deep nonnegative matrix factorization}.
\newblock {\em Big Data Mining and Analytics}, 3(1):13--28, 2019.

\bibitem{yu2018learning}
Jinshi Yu, Guoxu Zhou, Andrzej Cichocki, and Shengli Xie.
\newblock {Learning the hierarchical parts of objects by deep non-smooth
  nonnegative matrix factorization}.
\newblock {\em IEEE Access}, 6:58096--58105, 2018.

\bibitem{yang2021orthogonal}
Mingming Yang and Songhua Xu.
\newblock {Orthogonal Nonnegative Matrix Factorization using a novel deep
  Autoencoder Network}.
\newblock {\em Knowledge-Based Systems}, 227:107236, 2021.

\bibitem{zhao2019deep}
Yang Zhao, Huiyang Wang, and Jihong Pei.
\newblock {Deep non-negative matrix factorization architecture based on
  underlying basis images learning}.
\newblock {\em IEEE Transactions on Pattern Analysis and Machine Intelligence},
  43(6):1897--1913, 2019.

\bibitem{zhan2021deep}
Zihao Zhan, Wen-Sheng Chen, Binbin Pan, and Bo~Chen.
\newblock {Deep Grouped Non-Negative Matrix Factorization Method for Image Data
  Representation}.
\newblock In {\em 2021 International Conference on Machine Learning and
  Cybernetics (ICMLC)}, pages 1--6. IEEE, 2021.

\bibitem{shu2021deep}
Zhen-qiu Shu, Xiao-jun Wu, Cong Hu, Cong-zhe You, and Hong-hui Fan.
\newblock {Deep semi-nonnegative matrix factorization with elastic preserving
  for data representation}.
\newblock {\em Multimedia Tools and Applications}, 80(2):1707--1724, 2021.

\bibitem{lee2020feature}
Seokjin Lee and Hee-Suk Pang.
\newblock {Feature extraction based on the non-negative matrix factorization of
  convolutional neural networks for monitoring domestic activity with acoustic
  signals}.
\newblock {\em IEEE Access}, 8:122384--122395, 2020.

\bibitem{zhang2016nonlinear}
Hui Zhang, Huaping Liu, Rui Song, and Fuchun Sun.
\newblock {Nonlinear Non-negative Matrix Factorization using Deep Learning}.
\newblock In {\em 2016 International Joint Conference on Neural Networks
  (IJCNN)}, pages 477--482. IEEE, 2016.

\bibitem{tong2019deep}
Ming Tong, Yiran Chen, Mengao Zhao, Haili Bu, and Shengnan Xi.
\newblock {A deep discriminative and robust nonnegative matrix factorization
  network method with soft label constraint}.
\newblock {\em Neural Computing and Applications}, 31(11):7447--7475, 2019.

\bibitem{shu2022robust}
Zhenqiu Shu, Qinghan Long, Luping Zhang, Zhengtao Yu, and Xiao-Jun Wu.
\newblock {Robust Graph Regularized NMF with Dissimilarity and Similarity
  Constraints for ScRNA-seq Data Clustering}.
\newblock {\em Journal of Chemical Information and Modeling},
  62(23):6271--6286, 2022.

\bibitem{dutta2022neural}
Prasun Dutta and Rajat~K De.
\newblock {A Neural Network Model for Matrix Factorization: Dimensionality
  Reduction}.
\newblock In {\em 2022 IEEE Asia-Pacific Conference on Computer Science and
  Data Engineering (CSDE)}, pages 1--6. IEEE, 2022.

\bibitem{dutta2022n}
Prasun Dutta and Rajat~K De.
\newblock {n$^2$MFn$^2$: Non-negative Matrix Factorization in A Single
  Deconstruction Single Reconstruction Neural Network Framework for
  Dimensionality Reduction}.
\newblock In {\em 2022 International Conference on High Performance Big Data
  and Intelligent Systems (HDIS)}, pages 79--84. IEEE, 2022.

\bibitem{duttaPAAdn3mf}
Prasun Dutta and Rajat~K De.
\newblock {DN3MF: Deep Neural Network for Non-negative Matrix Factorization
  towards Low Rank Approximation}.
\newblock {\em Pattern Analysis and Applications}.

\bibitem{prasun2023MDSR}
Prasun Dutta and Rajat~K. De.
\newblock {MDSR-NMF: Multiple deconstruction single reconstruction deep neural
  network model for non-negative matrix factorization}.
\newblock {\em Network: Computation in Neural Systems}, 34(4):306--342, 2023.
\newblock PMID: 37818635.

\bibitem{kim2003subsystem}
Philip~M Kim and Bruce Tidor.
\newblock {Subsystem Identification Through Dimensionality Reduction of
  Large-Scale Gene Expression Data}.
\newblock {\em Genome Research}, 13(7):1706--1718, 2003.

\bibitem{pedregosa2011scikit}
Fabian Pedregosa, Ga{\"e}l Varoquaux, Alexandre Gramfort, Vincent Michel,
  Bertrand Thirion, Olivier Grisel, Mathieu Blondel, Peter Prettenhofer, Ron
  Weiss, Vincent Dubourg, et~al.
\newblock {Scikit-learn: Machine learning in Python}.
\newblock {\em The Journal of Machine Learning Research}, 12:2825--2830, 2011.

\bibitem{kingma2014adam}
Diederik~P Kingma and Jimmy Ba.
\newblock {Adam: A method for stochastic optimization}.
\newblock {\em arXiv preprint arXiv:1412.6980}, 2014.

\bibitem{mesejo2016computer}
Pablo Mesejo, Daniel Pizarro, Armand Abergel, Olivier Rouquette, Sylvain
  Beorchia, Laurent Poincloux, and Adrien Bartoli.
\newblock {Computer-aided classification of gastrointestinal lesions in regular
  colonoscopy}.
\newblock {\em IEEE Transactions on Medical Imaging}, 35(9):2051--2063, 2016.

\bibitem{fernandes2015proactive}
Kelwin Fernandes, Pedro Vinagre, and Paulo Cortez.
\newblock {A proactive intelligent decision support system for predicting the
  popularity of online news}.
\newblock In {\em Proceedings of the Portuguese Conference on Artificial
  Intelligence}, pages 535--546. Springer, 2015.

\bibitem{sakar2019comparative}
C~Okan Sakar, Gorkem Serbes, Aysegul Gunduz, Hunkar~C Tunc, Hatice Nizam,
  Betul~Erdogdu Sakar, Melih Tutuncu, Tarkan Aydin, M~Erdem Isenkul, and Hulya
  Apaydin.
\newblock {A comparative analysis of speech signal processing algorithms for
  Parkinson’s disease classification and the use of the tunable Q-factor
  wavelet transform}.
\newblock {\em Applied Soft Computing, Elsevier}, 74:255--263, 2019.

\bibitem{cortez2008using}
Paulo Cortez and Alice Maria~Gon{\c{c}}alves Silva.
\newblock {Using data mining to predict secondary school student performance}.
\newblock In {\em the Proceedings of 5th Annual Future Business Technology
  Conference, Porto, Portugal}, pages 5--12. The European Multidisciplinary
  Society for Modelling and Simulation Technology, The European Technology
  Institute Bvba (EUROSIS-ETI), 2008.

\bibitem{Dua:2019}
Dheeru Dua and Casey Graff.
\newblock {UCI} machine learning repository, 2019.

\bibitem{jiang2018direct}
Binyan Jiang, Xiangyu Wang, and Chenlei Leng.
\newblock {A direct approach for sparse quadratic discriminant analysis}.
\newblock {\em The Journal of Machine Learning Research}, 19(1):1098--1134,
  2018.

\bibitem{laugel2017inverse}
Thibault Laugel, Marie-Jeanne Lesot, Christophe Marsala, Xavier Renard, and
  Marcin Detyniecki.
\newblock {Inverse classification for comparison-based interpretability in
  machine learning}.
\newblock {\em arXiv preprint arXiv:1712.08443}, 2017.

\bibitem{laugel2018comparison}
Thibault Laugel, Marie-Jeanne Lesot, Christophe Marsala, Xavier Renard, and
  Marcin Detyniecki.
\newblock {Comparison-based inverse classification for interpretability in
  machine learning}.
\newblock In {\em Proceedings of the International Conference on Information
  Processing and Management of Uncertainty in Knowledge-Based Systems}, pages
  100--111. Springer, 2018.

\bibitem{harper2015movielens}
F~Maxwell Harper and Joseph~A Konstan.
\newblock {The movielens datasets: History and context}.
\newblock {\em Acm Transactions on Interactive Intelligent Systems (TIIS)},
  5(4):1--19, 2015.

\bibitem{glorot2010understanding}
Xavier Glorot and Yoshua Bengio.
\newblock {Understanding the difficulty of training deep feedforward neural
  networks}.
\newblock In {\em Proceedings of the thirteenth international conference on
  artificial intelligence and statistics}, pages 249--256. JMLR Workshop and
  Conference Proceedings, 2010.

\bibitem{van2009Learning}
Laurens van~der Maaten.
\newblock {Learning a Parametric Embedding by Preserving Local Structure}.
\newblock In David van Dyk and Max Welling, editors, {\em Proceedings of the
  Twelth International Conference on Artificial Intelligence and Statistics},
  volume~5 of {\em Proceedings of Machine Learning Research}, pages 384--391,
  Hilton Clearwater Beach Resort, Clearwater Beach, Florida USA, 16--18 Apr
  2009. PMLR.

\bibitem{Venna2001Neighborhood}
Jarkko Venna and Samuel Kaski.
\newblock {Neighborhood Preservation in Nonlinear Projection Methods: An
  Experimental Study}.
\newblock In Georg Dorffner, Horst Bischof, and Kurt" Hornik, editors, {\em
  Artificial Neural Networks --- ICANN 2001}, pages 485--491, Berlin,
  Heidelberg, 2001. Springer Berlin Heidelberg.

\end{thebibliography}
\end{document}